\documentclass[preprint,10pt,authoryear, twocolumn, 3p]{elsarticle}

\usepackage[english]{babel}
\usepackage[utf8]{inputenc}
\usepackage{amsmath}
\usepackage{graphicx}
\usepackage[colorinlistoftodos]{todonotes}
\usepackage{comment}
\usepackage{color,soul} 
\usepackage{makecell}
\usepackage{pifont}
\usepackage{ragged2e}
\usepackage{tabularx} 
\usepackage{booktabs}
\usepackage{etoolbox}

\journal{Neural Networks}

\begin{document}

\bibliographystyle{elsarticle-harv}

\begin{frontmatter}

\title{\textbf{A Neurocomputational Account of Flexible Goal-directed Cognition and Consciousness: The Goal-Aligning Representation Internal Manipulation Theory (GARIM)}}

\author[label1]{Giovanni Granato\corref{cor1}}
\address[label1]{Laboratory of Embodied Natural and Artificial Intelligence, Institute of Cognitive Sciences and Technologies, National Research Council of Italy, Rome, Italy}

\ead{giovanni.granato@istc.cnr.it}

\author[label1]{Gianluca Baldassarre}

\cortext[cor1]{I am corresponding author}

\ead{gianluca.baldassarre@istc.cnr.it}



\begin{abstract}
Goal-directed manipulation of representations is a key element of human flexible behaviour, while consciousness is often related to several aspects of higher-order cognition and human flexibility. Currently these two phenomena are only partially integrated (e.g., see Neurorepresentationalism) and this (a) limits our understanding of neuro-computational processes that lead conscious states to produce flexible goal-directed behaviours, 
(b) prevents a computational formalisation of conscious goal-directed manipulations of representations occurring in the brain, and (c) inhibits the exploitation of this knowledge for modelling and technological purposes. Addressing these issues, here we extend our `three-component theory of flexible cognition' by proposing the `Goal-Aligning Representations Internal Manipulation' (GARIM) theory of conscious and flexible goal-directed cognition. The central idea of the theory is that conscious states support the active manipulation of goal-relevant internal representations (e.g., of world states, objects, and action sequences) to make them more aligned with the pursued goals. This leads to the generation of the  knowledge which is necessary to face novel situations/goals, thus increasing the flexibility of goal-directed behaviours. The GARIM theory integrates key aspects of the main theories of consciousness into the functional neuro-computational framework of goal-directed behaviour. Moreover, it takes into account the subjective sensation of agency that accompanies conscious goal-directed processes (`GARIM agency'). The proposal has also implications for experimental studies on consciousness and clinical aspects of conscious goal-directed behaviour. Finally, the GARIM theory can benefit technological fields such as autonomous robotics and machine learning (e.g., the manipulation process may describe the operations performed by systems based on transformers).
\end{abstract}

\end{frontmatter}

\footnotemark{ The authors have equally contributed to the paper.}

\section{Introduction} 

\label{Sec:Introduction}

Goal-directed processes are at the basis of human flexible behaviour. We recently proposed and validated the `three-component theory of flexible cognition'
\citep{granato2020computational, granato2021internal, granato2022computational, granato2023flexible}, highlighting that a goal-directed top-down manipulation of representations is at the basis of cognitive flexibility. 
Although the our theory successfully describes the neuro-cognitive processes at the basis of cognitive flexibility, it does not focus on the higher-order processes underpinning flexible cognition (e.g., Planning and Problem Solving) nor on the role of consciousness in such processes.

Consciousness is a vastly debated concept and many theories formalise its key features \citep{seth2022theories}, 
which focus on different aspects such as
the integration of information 
\citep{Tononi2008Consciousnessasintegratedinformationaprovisionalmanifesto, TononiBolyMassiminiKoch2016Integratedinformationtheoryfromconsciousnesstoitsphysicalsubstrate, koch2016neural, Tononi2004AnInformationIntegrationTheoryofConsciousness};
the hierarchical convergence and divergence zones elaborating cognitive/emotional brain information \citep{Damasio1989Thebrainbindsentitiesandeventsbymultiregionalactivationfromconvergencezones, MeyerDamasio2009Convergenceanddivergenceinaneuralarchitectureforrecognitionandmemory,DamasioMeyer2009ConsciousnessanOverviewofthePhenomenonandofItsPossibleNeuralBasis};
the selection of relevant information into a central workspace and its `broadcasting' to peripheral areas
\citep{Baars1997, Baars2003, Baars2005, Baars2013};
the  top-down activation of multiple hierarchical brain systems by the frontoparietal system
\citep{Dehaene1998, Dehaene2001, DehaeneChangeux2011Experimentalandtheoreticalapproachestoconsciousprocessing}; 
the difference between first-order and higher-order representations \citep{brown2019understanding,cleeremans2011radical}; the coordination of effective brain-body-environment sensorimotor interactions \citep{OReganNoee2001Asensorimotoraccountofvisionandvisualconsciousness,o2005sensory};
the emergence of multi-modal/multi-level representations that subserve goal-directed behaviours \citep{Pennartz2015TheBrainsRepresentationalPoweronConsciousnessandtheIntegrationofModalities, pennartz2018consciousness, pennartz2022neurorepresentationalism};
the dynamic loops that generate and adjust predictions based on an inferential process \citep{clark2013whatever, hohwy2020predictive, friston2018self}.
These theories consider some key aspects that are commonly related to goal-directed cognition such as information hierarchies, top-down information selection, sensory-motor interactions, multi-modal integration. However, with the exception of neurorepresentationalism \citep{pennartz2022neurorepresentationalism}, most theories capture only few elements of goal-directed cognition.

This general poor integration leads to several scientific and technological issues.
First, it limits the emergence of integrated frameworks that establish the relationships between consciousness, goal-directed cognition and flexible behaviour. In particular, it limits the understanding of the neuro-computational processes that lead conscious states to produce a more flexible goal-directed behaviour.
Second, most theories of consciousness do not analyse the fine computational processes that support a conscious and goal-directed manipulation of information. Although some theories have led to computational models of consciousness (e.g., \citealp{dehaene1998neuronal, pasquali2010know, Tononi2008Consciousnessasintegratedinformationaprovisionalmanifesto}), there is not yet a clear description of the system-level manipulations of information that occur during conscious goal-directed processing.
Third, these limitations inhibit an effective and broad exploitation of these models and formalisations for technological scopes. 
Although the emergence of interdisciplinary fields such as \textit{machine consciousness} \citep{reggia2013rise} and consciousness-inspired machine learning (e.g., \citealp{bengio2017consciousness}), artificial intelligence (AI) and autonomous robot systems have till recently shown rigid behaviours and processes, in particular have failed to face novel conditions or goals \citep{baldassarre2020goal, HassabisKumaranSummerfieldBotvinick2017NeuroscienceInspiredArtificialIntelligence,LakeUllmanTenenbaumGershman2017Buildingmachinesthatlearnandthinklikepeople_TargetPaper,BubeckChandrasekaranEldanGehrkeHorvitzKamarLeeLeeLiLundbergNoriPalangiRibeiroZhang2023SparksOfArtificialGeneralIntelligenceEarlyExperimentsWithGPT4}.
Only recently some AI systems have exhibited examples of flexible general-purpose cognition \citep{LiLiXiongHoi2022BlipBootstrappingLanguageImagePreTrainingForUnifiedVisionLanguageUnderstandingAndGeneration,DriessXiaSajjadiLynchChowdheryIchterWahidTompsonVuongYuHuangChebotarSermanetDuckworthLevineVanhouckeHausmanToussaintGreffZengMordatchFlorence2023PaLMEAnEmbodiedMultimodalLanguageModel,ParkOBrienCaiMorrisLiangBernstein2023GenerativeAgentsInteractiveSimulacraOfHumanBehavior}.
Interestingly, these models rely on new algorithms (`transformers') that have notable links with the manipulation of internal representations. However, the functioning of these systems is still poorly understood \citep{2021DoesVisionandLanguagePretrainingImproveLexicalGrounding,AbdouKulmizevHershcovichFrankPavlickSoegaard2021CanLanguageModelsEncodePerceptualStructureWithoutGroundingACaseStudyInColor,AgueerayArcas2022DoLargeLanguageModelsUnderstandUs,SrivastavaEtAl2022BeyondTheImitationGameQuantifyingAndExtrapolatingTheCapabilitiesOfLanguageModels}. 
Overall, these  limitations negatively impact both scientific and technological advancements, which would benefit from a adequate integration between studies on goal-directed behaviour and higher-order aspects of consciousness.

Addressing these issues, we extend here our `three-component theory of flexible cognition' to higher-order aspects of goal-directed cognition and consciousness, thus introducing the \textit{Goal-Aligning Representation Internal Manipulation} (GARIM) theory of higher-order cognition and consciousness.
The core idea of the GARIM theory is that conscious processes enhance the flexibility of goal-directed behaviour by supporting the manipulation of goal-relevant internal representations (e.g., of world states, objects, and action sequences). These manipulations generate the knowledge that the agent lacks to improve the \textit{alignment} of such representations with the target goals, especially when these are new or are pursued in novel conditions. Therefore, a higher goal-oriented `alignment' makes active representations more likely to generate successfully goal-oriented actions.

The GARIM theory is based on five key features: 
(1) an adaptive function of consciousness; 
(2) specific representations at the basis of conscious and flexible goal-directed behaviours (Goal-based Integrated Neural Patterns; GINPs); 
(3) four interacting neurofunctional systems supported by cortical networks and basal ganglia (hierarchical perceptual working-memories, abstract working memory, internal manipulator, motivational systems) at the basis of conscious goal-directed manipulations of GINPs;
(4) four macro-classes of representation manipulations which generate the goal-oriented missing knowledge (`abstraction', `specification', 
`decomposition', `composition');
(5) the emergence of a subjective sensation of agency related to the representation manipulations (`GARIM agency').
This last feature contributes to differentiate the concept of agency that emerges during a goal-directed behaviour from those related to consciousness and conscious states.
Although the GARIM theory introduces many considerations about low-order cognition (e.g., motivational and emotional aspects of conscious flexible behaviour), it mainly focuses on higher-order aspects of conscious cognition (e.g., top-down attention and planning).
This is consistent with the theory functionalist approach and technological implications, as we show by linking it to concepts on higher-order human cognition and artificial intelligence (see section `GARIM theory and intelligence'). 
For the same reason, here we focus on meso-scale aspects of the brain (e.g. interactions of brain macro-systems and broad neural representations) rather then on neuron-level aspects of it. 
This level of analysis is suitable for addressing the functions and computations targeted here, thus both for studying human cognition and for the development of artificial intelligent systems and robots. 

The GARIM theory primarily aims to clarify the neuro-computational processes that actively lead to more flexible goal-directed behaviours. In this respect, the theory can be conceived as a \textit{neuro-computational framework of conscious and flexible goal-directed cognition.}
Moreover, the theory gives four specific theoretical contributions. 
First, it clarifies some aspects of subjective experience and agency.
In particular, it proposes the concept of the GARIM agency to explain the different subjective experiences that accompany conscious states.
Second, the theory contributes to specify the neurocomputational mechanisms underlying the main theories on consciousness.
This allows the integration of those mechanisms within a common functional and computational framework that pivots on goal-directed processes.
Third, the theory generates insights for the experimental and clinical fields related to conscious states and goal-directed behaviours.
In particular, the theory is shown to be compatible with relevant experimental predictions of other theories on consciousness.
Moreover, the theory gives indications for building new experimental paradigms for testing consciousness.
Last, the theory offers an interpretation of the relationships between certain clinical impairments and conscious goal-directed behaviours.
Fourth, the theory provides some insights that could be useful for building new computational models, ML-based systems, and robotic architectures. Also, it can be useful to analyze existing ones.
In this regard, computational models can operationalise the theory, allowing it to be corroborated with specific empirical data and more detailed comparisons with other theories.
On the other hand, indications towards ML and robotics could improve the goal-directed flexibility of current systems and their interpretations \citep{baldassarre2020goal, HassabisKumaranSummerfieldBotvinick2017NeuroscienceInspiredArtificialIntelligence,LakeUllmanTenenbaumGershman2017Buildingmachinesthatlearnandthinklikepeople_TargetPaper}.

\begin{figure}[htb!]
  \centering
   \includegraphics[width = 0.50\textwidth]{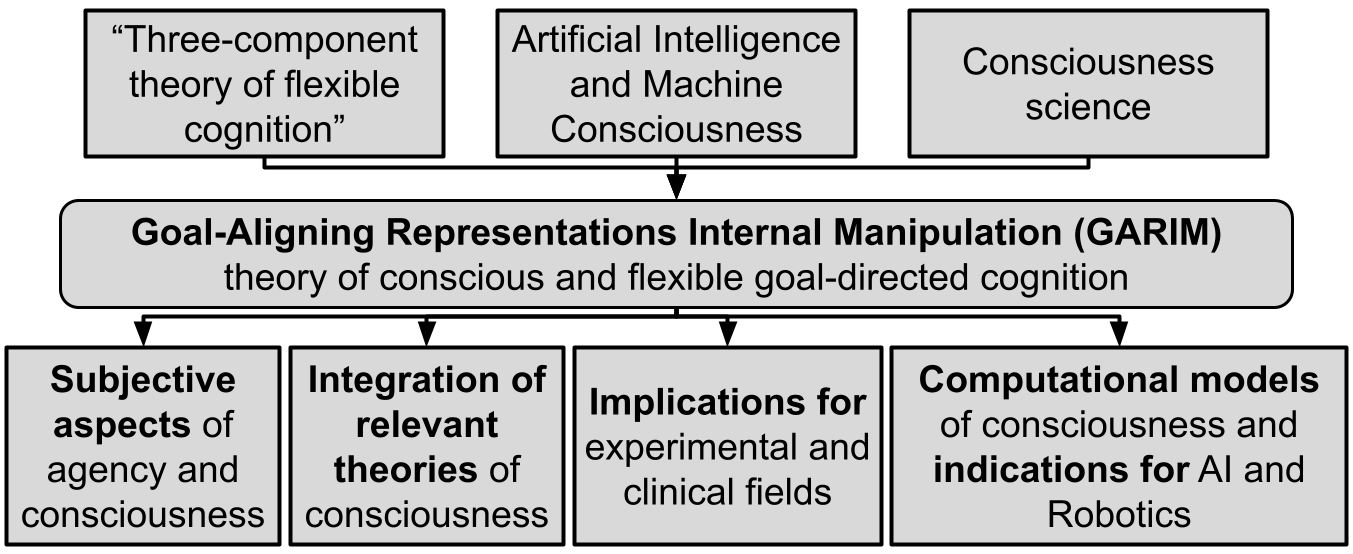}
  \vspace*{+1mm}
  \caption{The schema shows the key fields leading to the development of the GARIM theory and its main contributions. This schema also represents the overall organisation of this work.
}
  \label{Figure:RIMPaperPert}
\end{figure}

Figure~\ref{Figure:RIMPaperPert} summarises the main contributions of the GARIM theory and the organisation of this work. 
We first describe the three-component theory of flexible cognition, highlighting its key features, limitations, and also related technological fields (in particular AI and robotics).
Building on such theory, we then introduce the key features of the GARIM theory.
Next, we compare our proposal with the major theories of consciousness, showing that key higher-order aspects of consciousness are captured by our integrated neuro-computational framework.
We then analyse the empirical implications of the theory by considering both experimental and clinical evidence.
Finally, we consider the implications for the design of new computational models, AI systems and robotic architectures.
%


\section{The three-component theory of flexible cognition: key features, limitations and related fields}

The three-component theory of flexible cognition formalises the neuro-cognitive processes that boost cognitive flexibility during goal-directed behaviours \citep{granato2021internal, granato2020computational}. 
In particular, the theory describes behaviours based on learnt action-outcome associations and goals \citep{BalleineOstlund2007StillattheChoicePointActionSelectionandInitiationinInstrumentalConditioning, ThillCaligioreBorghiZiemkeBaldassarre2013TheoriesandcomputationalmodelsofaffordanceandmirrorsystemsAnintegrativereview,passingham2012neurobiology,tsujimoto2011frontal}. In this context, `goals' are defined as representations of desirable future states that (a) can be stored and internally re-activated and (b) can lead to select actions directed to achieve them. 
Moreover, the theory operationalises `cognitive flexibility', an executive function that is defined as the capacity of switching between different representations depending on external and internal feedback \citep{Diamond2013Executivefunctions}.

The main idea of the three-component theory is that \textit{flexible cognition depends on the top-down goal-directed manipulation of representations}. These manipulated representations better support sensory–motor interactions with the environment, thereby boosting cognitive flexibility and goal-directed behaviour.

The theory proposes that three main elements are at the basis of flexible goal-directed behaviour (Table~\ref{Table:ThreeComponentsTheory_KeyFeatures}).

\begin{table}[htb!]
\centering
\resizebox{0.5\textwidth}{!}{
\begin{tabular}{c  c}
\hline
\textbf{Main elements} & \textbf{Explanation} \\
\hline
\textbf{\thead{Three key \\ components}}
& \thead{\\   Three neuro-functional systems    \\   support the goal-directed top-down    \\   manipulation of representations:    \\\\
  $\cdot$ Executive Working-Memory    \\
  $\cdot$ Hierarchical perceptual systems    \\
  $\cdot$ Top-down manipulator   } \\\\
\hline
\\
\textbf{\thead{First-order and second-order \\ representations/manipulations}} & \thead{  Two levels of representations    \\   and manipulations:    \\\\
  $\cdot$ First-order (e.g., perceptions    \\ 
  and selective attention)    \\
  $\cdot$ Second-order (e.g., abstract    \\   goals and inner-speech) }   \\\\
\hline
\\
\textbf{\thead{Embodied \\ sensory-motor loops}} & \thead{  Representation manipulations and    \\   sensory-motor embodied loops    \\   support goal-directed behaviours   } \\\\
\hline
\end{tabular}
}
\vspace*{+3mm}
\caption{The table shows the main elements of the three-component theory of flexible goal-directed cognition.}
\label{Table:ThreeComponentsTheory_KeyFeatures}
\end{table}

First, three neuro-functional systems support the goal-directed representation manipulation. An \textit{Executive Working-Memory}, supported by the prefrontal cortices (PFC) and frontal-striatal loops, stores the goals/sub-goals \citep{hartley2000locating, braver2002role}. \textit{Multiple hierarchical perceptual systems}, supported by cortical perceptual pathways, extract and store the goal-related representations \citep{RizzolattiMatelli2003Twodifferentstreamsformthedorsalvisualsystemanatomyandfunctions, gazzaley2012top, raffone2014interplay}. \textit{A top-down manipulator}, supported by frontal-parietal cortical system and basal ganglia-thalamo-cortical loops, applies a goal-directed manipulation of the stored representations at different stages of hierarchical systems \citep{vossel2014dorsal, parks2013brain, redgrave1999basal,seger2008basal, chelazzi2013rewards, pessoa2015multiple}. Note that further systems can act as a top-down manipulator (e.g., the inner-speech system; \citealp{granato2020computational, granato2022computational, granato2023flexible}).

Second, the theory distinguishes between `first-order' and `second-order' representations and manipulations \citep{granato2020computational}. The first term refers to perceptual representations and their manipulations (e.g., visual selective attentional processes). The second term refers to abstract/amodal representations (e.g., goals/sub-goals, actions) and their manipulations (e.g., splitting of goals into sub-goals). Both refer to a self-directed form of manipulation at different levels of abstraction.

Third, a synergistic interplay between the goal-directed representation manipulation and embodied sensory-motor loops is central to express flexible goal-directed behaviour. In particular, goal-directed behaviour is supported by multiple manipulations of internal representations and the external world.

The three-component theory is validated trough an integrated theory-driven/data-driven computational approach. In particular, we developed a neuro-inspired computational model based on this theory and we tested it with a neuropsychological test of cognitive flexibility (for further details see the section `Towards computational models of the GARIM theory'). The computational model has reproduced the behavioural data obtained from various cohorts of human participants, both in healthy and clinical conditions \citep{granato2021internal, granato2020computational, granato2022computational, granato2023flexible}.

\subsection{Beyond the three-component theory}

\label{Sec:Beyond_TCT}

Although the three-component theory has received an experimental and computational validation, it shows limitations. 
First, although executive functions and goal-directed behaviours are often linked to explicit cognition and consciousness, the three-component theory does not take into account the role of conscious processes in flexible cognition. Despite this limitation, the theory formalises some key neurocognitive processes that are central to the emergence of conscious states, such as information hierarchies \citep{damasio2009consciousness}, top-down information selection \citep{Dehaene2011}, sensory-motor interactions \citep{o2005sensory}, and first-order/second-order representations \citep{brown2019understanding}.
Second, the theory does not formalise monitoring processes and related higher-order goal-directed behaviours such as `planning', defined as a flexible assembling of new action sequences to accomplish goals \citep{pfeiffer2013hippocampal,delatour2000functional}, and `problem solving', defined as a planning process that involves partial knowledge \citep{newman2003frontal}. These limitations are due to the fact that the theory does not take into consideration three key aspects of higher-order cognition: world models, `motivations', and `emotions'. The following paragraphs briefly articulate these concepts and their role in goal-directed behaviour.

\paragraph{World models}
The theory takes into account the concept of `goals', formalised as stored working-memory representations  that change over time, but it does not consider the concept of \textit{world model} \citep{MarsSalletRushworthYeung2011NeuralBasisofMotivationalandCognitiveControl,passingham2012neurobiology,fuster2015past}. World models are representations of the spatiotemporal dynamics of the environment, integrating knowledge on the evolution of the physical environment and the effects that actions cause on it   \citep{soltani2022computational}. World models support planning and problem solving as they allow the agent to internally simulate the dynamic transitions of the environment, from a starting state to final goals. Therefore, by integrating percept, goals and spatiotemporal simulations, they support monitoring/ conflict resolution processes and drive goal-directed plans \citep{huddy2023loss, powers2016perceptual}.
World model representations and processes are mostly supported by PFC systems and their loops with sub-cortical structures (in particular basal ganglia-thalamus and hippocampal systems; \citealp{HoukDavidsBeiser1995,Fuster2008Theprefrontalcortex,tang2021multiple,PataiSpiers2021TheVersatileWayfinderPrefrontalContributionstoSpatialNavigation,HaszRedish2020SpatialEncodinginDorsomedialPrefrontalCortexandHippocampusIsRelatedduringDeliberation}). 

\paragraph{Motivations and emotions}
Although the three-component theory implicitly assumes that motivational signals act as feedback to change the stored goals, it does not explicitly examine the role of \textit{motivations} and \textit{emotions} at the basis of flexible cognition.
Motivations support the formation and reactivation of goal representations during goal-directed behaviours, guide learning, contribute to select behaviors to perform and energise them \citep{MarsSalletRushworthYeung2011NeuralBasisofMotivationalandCognitiveControl}.
Motivations can be divided into extrinsic/physiological motivations (e.g., for safety,  water, and food), social motivations (e.g., for belonging to a group), and intrinsic motivations (e.g., novelty, surprise, competence improvement, serving knowledge and skills acquisition) \citep{Panksepp1998AffectiveNeurosciencetheFoundationsofHumanandAnimalEmotions,GangestadGrebe2017,RyanDeci2000Selfdeterminationtheoryandthefacilitationofintrinsicmotivationsocialdevelopmentandwellbeing,mirolli2010roles}.
The role of emotions is still a debated topic \citep{Scherer2005WhatAreEmotionsandHowCanTheyBeMeasured,Cabanac2002WhatIsEmotion}. Many studies suggest that they promote the production of adaptive behaviours (e.g., engagement, avoidance, and social communication; \citealp{Panksepp1998AffectiveNeurosciencetheFoundationsofHumanandAnimalEmotions,EkmanDavidson1994TheNatureofEmotionFundamentalQuestions,Damasio1998EmotioninthePerspectiveofanIntegratedNervousSystem}). In general, emotions predispose the body and brain to get into specific adaptive overall modes of functioning.
A bulk of studies investigated neural correlates of motivations and emotions \citep{Panksepp1998AffectiveNeurosciencetheFoundationsofHumanandAnimalEmotions,Schultz2002Gettingformalwith,Frith2007,Amaral2002,Rolls2004,lisman2005hippocampal,Paus2001Primateanteriorcingulatecortexwheremotorcontroldriveandcognitioninterface,o2013dissociable,RibasFernandesSolwayDiukMcGuireBartoNivBotvinick2011Aneuralsignatureofhierarchicalreinforcementlearning}. These studies show that they are supported by many interacting sub-cortical structures (e.g., hypothalamus, amygdala, insula, hippocampus) and cortical structures (e.g., medial/temporal cortex, orbitofrontal cortex, anterior cingulate cortex, prefrontal cortex). 


%
Note that the terms `motivations' and `emotions' can refer to a global brain state (motivational state; e.g. schizophrenia patients show a low motivational state defined `apathy'; \citealp{bortolon2018apathy}) or localised events (motivational signal; e.g., dopamine bursts support information selection and learning; \citealp{berke2018does}). 
The two elements interact (e.g., apathy in schizophrenic patients lead to an inefficient information selection; \citealp{bortolon2018apathy}) and contribute to shape conscious states and goal-directed cognition. However, in this work we particularly refers to the second function (motivational/emotional signals at the basis of the information selection components).


\subsubsection{Contributions of Artificial Intelligence and Robotics}

AI and autonomous robotics give important inputs to extend the three-component theory towards the GARIM theory. In particular, they both contribute to the investigation of brain and cognition by supporting their computational modelling (see section `Towards computational models of the GARIM theory'). At the same time, as discussed in section `Towards AI systems and robotics architectures inspired by the GARIM theory', they might benefit from the scientific knowledge on brain and cognition to build more efficient and flexible intelligent machines \citep{BaldassarreSantucciCartoniCaligiore2017ThearchitecturechallengeFutureartificialintelligencesystemswillrequiresophisticatedarchitecturesandknowledgeofthebrainmightguidetheirconstruction, baldassarre2020goal}. 

\paragraph{Artificial intelligence, machine learning, and neural networks}

Goal-oriented processes have always played a central role in artificial intelligence \citep{RussellNorvig2016ArtificialIntelligenceAModernApproach}. In particular, AI has always attributed to human intelligence the primarily role of accomplishing goals through the search of the most suitable starting state-goal sequence \citep{simon1975functional}. 

Interestingly, the type of encoding of the representations of planning elements strongly affected the evolution of the field.
In particular, initial problem solving systems used `atomic' representations (i.e., distinct symbols for states and actions), which made the action sequence search inefficient due to combinatorial explosion. 
Later, studies on \textit{planning} \citep{RussellNorvig2016ArtificialIntelligenceAModernApproach} `factorised' the representations of states into elements (e.g., `objects') and relations between elements (e.g., `being part of', `being on'). This change reduced the computational costs at the basis of the action-sequence search.

In parallel, `connectionist approaches' based on neural-networks proposed alternative systems based on `sub-symbolic representations', namely representations of features encoded in neural patterns \citep{McClellandPDPResearchGroup1986ParallelDistributedProcessingExplorationsintheMicrostructureofCognition}. Neural networks, initially used in machine learning to implement `reactive processes', have been recently used to implement goal-directed processes such as planning (e.g., \citealp{rehder2018pedestrian, WayneHungAmosMirzaAhujaGrabskaBarwinskaRaeMirowskiLeiboSantoroGemiciReynoldsHarleyAbramsonMohamedRezendeSaxtonCainHillierSilverKavukcuogluBotvinickHassabisLillicrap2018UnsupervisedPredictiveMemoryinaGoalDirectedAgent}). Furthermore, recent research has proposed that `deep neural networks' could model key processes underlying human consciousness \citep{bengio2017consciousness}.

Overall, artificial intelligence studies highlight how higher-order cognition can benefit of disentangled and factored representations that can (a) be combined into new ones and (b) represent interdependencies among their sub-parts.

\paragraph{Machine consciousness: key elements at the basis of higher-order cognition and consciousness}

Machine consciousness (MC) is a research field aiming to define the key elements that artificial-intelligence and robotic systems should have to exhibit a certain level of consciousness \citep{aleksander1995artificial, gamez2008progress}.
MC adopts both scientific and technological approaches to accomplish this objective \citep{reggia2013rise}.
The scientific approach aims to develop and validate computational models built on the basis of the main theories of consciousness.
The technological approach aims to integrate elements of consciousness into AI and robotic systems to improve their flexibility and adaptability.
\cite{AleksanderDunmall2003AxiomsandTestsforthePresenceofMinimalConsciousnessinAgentsIPreamble} proposes fives `axioms', stating which fundamental capabilities an intelligent system should have to exhibit a minimal level of consciousness:`depiction' (i.e. the capacity to represent elements of
the world), imagination, attention, planning, emotions.
On the other hand, \citet{gamez2008progress} proposes that MC systems can be grouped into four classes based on their `consciousness simulation level'.
A first class (MC1) involves the systems that exhibit a `conscious like' external behaviour, such as AIs that exhibit human-level competence in playing complex games \citep{ferrucci2012introduction, lewis2012game}.
A second class (MC2) encompasses systems that are generally inspired by theories of consciousness and show internal `cognitive processes' similar to those of conscious agents (e.g., attentional processes, motivation, world models; \citealp{kugele2021learning, franklin2012global, holland2007strongly, marques2009architectures, jantsch2010distributed}). 
The third class (MC3) involves systems that are inspired by theories of consciousness and show brain-inspired architectures \citep{dehaene2003neuronal, gamez2010information}. 
The fourth and final class (MC4) encompasses systems able to engage in phenomenological forms of conscious subjective experience. 
There is a hot debate regarding the implementation of this kind of artificial systems \citep{carter2018conscious, reggia2013rise}, but for now no artificial system seems to be able to undergo a human-like conscious internal experience.
At last, a relevant review \citep{reggia2013rise} highlights that the proposals of MC can be categorised in five key classes, built on specific core principles:
(1) internal models of the agent itself (self-modelling);
(2) information broadcasting;
(3) higher-order representations;
(4) attention processes;
(5) information integration.

These frameworks (the fundamental axioms, the levels of simulation, and the main implementation principles) support the formalisation of higher-order cognition and consciousness. All these elements have been taken into account to develop the GARIM theory.


\section{The Goal-Aligning Representation Internal Manipulation theory}
\label{Section:RIM_Main}

This section presents the five major elements of the
GARIM theory (see Table~\ref{Table:GARIMTheory_KeyFeatures}):
(a) the adaptive function that consciousness plays in goal-directed processes;
(b) specific neural patterns that form conscious goal-related representations; 
(c) four anatomo-functional macro-systems that support the manipulation of representations;
(d) four classes of computational operations that describe such manipulations;
(e) an explanation, based on the concept of GARIM agency, that links agency and subjective conscious experience. 
The following sections present these  elements in detail.

\begin{table}[htb!]
\centering
\resizebox{0.5 \textwidth}{!}{
\begin{tabular}{c  c}
\hline
\textbf{Main elements} & \textbf{Explanation} \\
\hline
\\
\textbf{\thead{Adaptive function \\ of consciousness}}
& \thead{  Consciousness improves flexibility: conscious  \\ 
  states and processes support representation    \\ 
   manipulations in order to increase their alignment   \\
  with pursued goals, thereby enhancing   \\
  goal-directed behaviours.  } \\\\
\hline
\\
\textbf{\thead{Goal-related \\  representations}} & \thead{   Conscious goal-directed behaviours are supported   \\ 
   by goal-directed integrated neural patterns (GINPs)  
\\   having two key dimensions, i.e. goal-relevance and   
\\   consciousness level.  } \\\\
\hline
\textbf{\thead{Four key \\ components}} & \thead{\\   Four neuro-functional systems support the goal-   \\    directed representation manipulation: abstract   \\
  working-memory, perceptual working-memory,   \\
  internal manipulator, motivational system.  } \\\\
\hline
\\
\textbf{\thead{GARIM \\ operations}} & \thead{  Goal-directed manipulations modify the GINPs   \\ and are subjectively experienced as intentionally \\ directed operations. They are divided in four  \\   classes of operations: abstraction,  specification   \\
  decomposition, composition.  } \\\\
\hline 
\\
\textbf{\thead{GARIM\\ agency}} & \thead{GARIM agency emerges during the execution \\ of conscious flexible goal-directed represen- \\ -tation manipulations. These generate a sub- \\ -jective internal reality featured by three key \\   elements:  self-models, emotional and  \\ perceptual vividness, and manipulation control.} \\\\
\hline
\end{tabular}
 }
\vspace*{+1mm}
\caption{The five major elements of the GARIM theory.}
\label{Table:GARIMTheory_KeyFeatures}
\end{table}

\subsection{The adaptive function of consciousness}
\label{sec:Adaptive_Function}

The GARIM theory postulates that \textit{the adaptive function of consciousness is the improvement of flexibility during the expression of goal-directed behaviours. 
In particular, conscious states enable agents to manipulate their internal representations (e.g., perceptions, thoughts and actions) in order to generate knowledge more aligned with the set goal; 
the higher goal-related alignment leads to more successful goal-oriented actions in familiar or novel situations, thereby enhancing goal-directed behaviours (e.g., decision-making, planning and problem-solving).}

This feature is consistent with the commonly recognised role of goal-directed cognition for human daily-life behaviours. Indeed, in case of new goals or situations habitual behaviours are often no longer suitable or efficient. Moreover, also in case of behaviours that successfully lead to specific sub-goals (e.g., an improvement of physical performance due to doping substances), they could show a  misalignment with higher-level goals (e.g., honestly get an Olympics medal). Our proposal specifies that representation manipulations generate more suitable knowledge (e.g., plans, objects' representations, sub-goals) to face these conditions.

Most theories of consciousness highlight that conscious processes have an adaptive role for human  behaviours (see section `Comparisons of the GARIM theory with other theories').
For example, global workspace theories link conscious states to information sharing and amplification at the basis of decision making. On the other hand, Predictive Processing theories link conscious processes to a continuous optimisation of inference/prediction mechanisms related to goal-directed behaviours. Importantly, the Neurorepresentationalism's framework explicitly links Consciousness to goal-directed behaviours. In particular, it suggests that conscious processes generate the best representations that serve goal-directed processes.

Our proposal is compatible with studies on goal-directed behaviour and consciousness. However, it proposes a specific focus on higher-order processes. For example, Neurorepresentationalism focuses on the emergence of representations that are subsequently exploited by goal-directed processes. Instead, the GARIM theory formalises mechanisms that operate and constitute goal-directed cognition (i.e., goal-directed representation manipulations).

\subsection{Goal-based Integrated Neural Patterns (GINPs): conscious representations at the basis of flexible goal-directed behaviour}

\begin{figure*}[htb!]
  \centering
   \includegraphics[width= \textwidth]{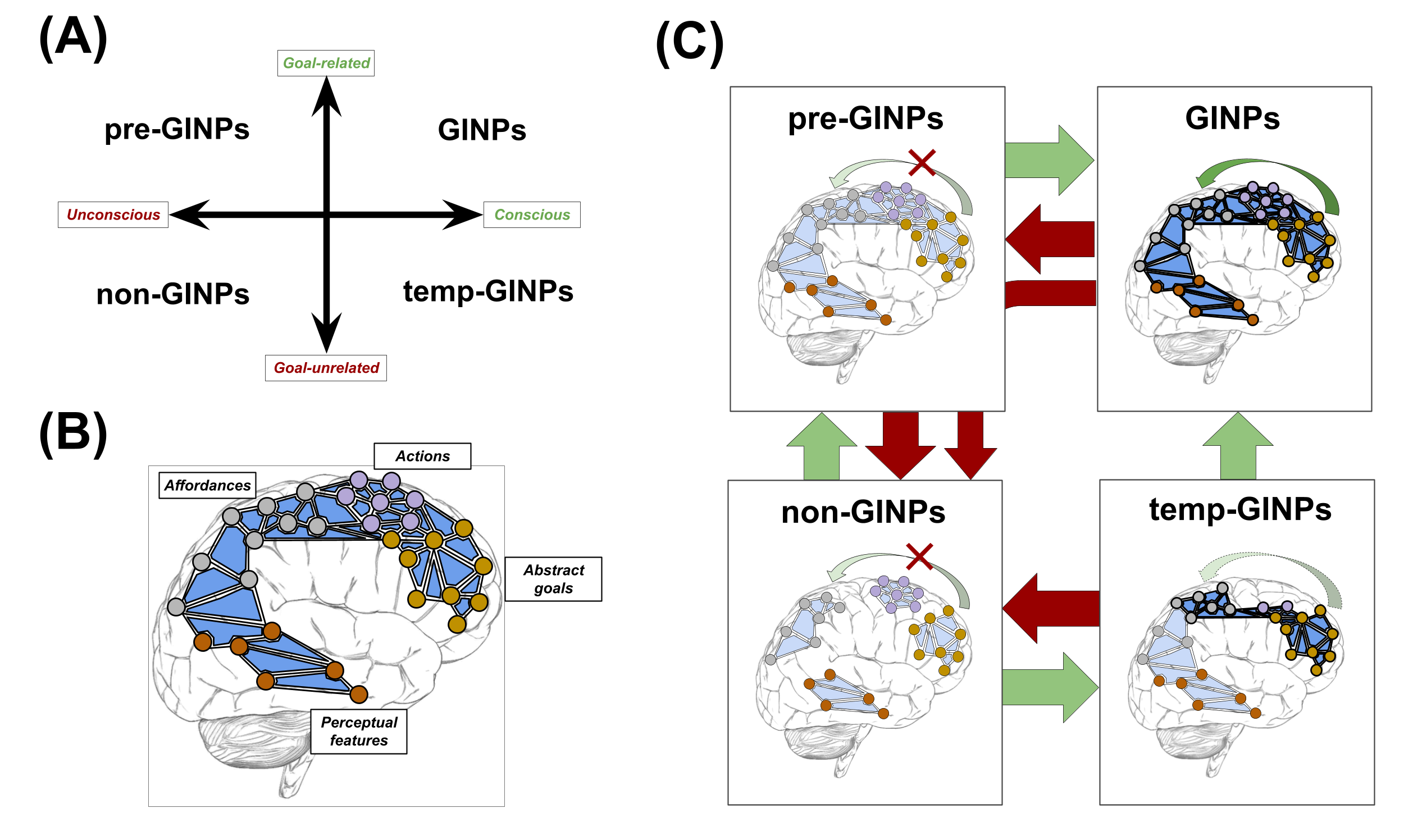}
  \vspace*{+1mm}
  \caption{
  (A) Different kinds of Goal-based Integrated Neural Patterns (GINPs), classified on the basis of their key dimensions: goal-relevance and conscious level.
  (B) Possible neural correlates of a GINP; the figure shows four sub-GINPs (orange, grey, violet, yellow), encoding different goal-relevant elements (e.g., perceptual features of objects, affordances, actions, abstract representations of goals). 
  (C) Possible sequential relationship between different kinds of GINPs; green arrows indicate an `unconscious-to-conscious' change (e.g., a pre-GINP that becomes a GINP) while red arrows indicate the inverse change. Note that a curved arrow, on top of each box, indicates the level of goal-relatedness/goal-based representation manipulation (e.g., absent, temporary, prolonged).}
  \label{Figure:GINPs_schema}
\end{figure*}

The GARIM theory describes brain states that support higher-order conscious cognition, leading to the expression of flexible goal-directed behaviour. To adequately represent the peculiarities of these states, we introduce the concept of
\textit{`Goal-based Integrated Neural Pattern'} (GINP; see Figure \ref{Figure:GINPs_schema}). 
Although we cannot present direct experimental evidence for GINPs, their existence is compatible with studies on goal-directed behaviour and other theories of consciousness (see the paragraph `GINPs and other definitions of conscious representations').

\paragraph{Definition, features and brain correlates}

We define a GINP as an active distributed neural representation that is characterised by two features: (a) \textit{consciousness level}: it is consciously perceived and thus intentionally manipulable; (b) \textit{goal-relevance}: it is functionally relevant for the pursued goals. GINPs are integrated representations that have a compound nature, in particular  are formed by sub-parts (`sub-GINPs'). These sub-GINPs encode different aspects of goal-directed contents (e.g., percepts, affordances, actions, goals). 

We hypothesise that GINPs are encoded at multiple levels by many structures in the brain hierarchies (see figure~\ref{Figure:GINPs_schema}). A GINP related to a specific goal (e.g., `patting a dog', encoded in the PFC) could be formed by sub-GINPs related to visual appearance (e.g., the dog aspect, encoded by visual areas), an overt sound (e.g., the barking, encoded by auditive areas), a inner-speech production (e.g., the word `dog', encoded by language areas), and possible related actions (e.g., `patting', encoded by motor areas). We expect that the strength of physical/functional connections between sub-GINPs (i.e., neural integration) varies depending on their consciousness level and goal-relevance. 

Dynamically, only one GINP can become conscious at a certain moment. This is consistent with the commonly accepted fact that only a limited representation can access consciousness at a time. However, the GINP continually evolves under the effect that representation manipulations have on its sub-GINPs.
The integration between the sub-GINPs could be supported by both physical connections and functional connectivity. Sub-GINPs can have such a low level of integration that they stop forming integrated representations. In this case, for simplicity, we still keep the `-GINP' word ending (see `Non-GINPs' in the next paragraph). 

\paragraph{Non-GINPs, Pre-GINPs, Temp-GINPs and GINPs: from unconscious to stably conscious representations, and vice versa}

We hypothesise that GINPs enable representation manipulations, and thus conscious and flexible goal-directed behaviours. However, on the basis of their consciousness level and goal-relevance, we define four kinds of representations that can emerge under the operation of consciousness processes. 

\textit{GINPs}: whole conscious representations that have a high level of goal-relevance and stability in time, and thus strongly affect goal-directed behaviour.
\textit{Temp-GINPs}: representations that have a low goal-relevance, but nevertheless temporarily access consciousness (e.g., salient stimuli such as unexpected skin pressure or distracting internal thoughts). They can acquire goal-relevance, thus becoming GINPs, or can be suppressed by top-down attention.
\textit{Pre-GINPs}: unconscious representations that have a high level of goal-relevance but do not have the support of top-down attention, thus remaining unconscious. They can be activated by background processes (e.g., priming) and can indirectly influence conscious representations.
\textit{Non-GINPs}: unconscious representations that have little or no goal-relevance, but are activated by external events or related internal active representations. Depending on their features, Non-GINPs could have a very low integration. Therefore, they could stop being `integrated global representations' and becoming `scattered local representations'. 

In our descriptions, we will generally refer to the whole brain representations (GINPs) but their sub-parts (sub-GINPs) will often inherit their consciousness-level/goal-relevance properties. For example, when we refer to a GINP as a `goal-relevant consciously perceived representation', we also imply its specific sub-GINPs are goal-relevant and accessible by consciousness processes.  
Instead, when we consider the different status of the currently active sub-GINPs, we refer to them separately.

Figure~\ref{Figure:GINPs_schema} (box C) shows that the four types of representations can exhibit a sequential dynamic relationship. For example, a non-GINP could progressively acquire goal-relevance until becomes a pre-GINP and then a GINP. Conversely, a GINP can be de-activated and become a pre-GINP; and in case it loses most of its goal-relevance it becomes a non-GINP. On the other hand, a non-GINP can temporally access consciousness with a low goal-relatedness, becoming a temp-GINP (e.g., representing an object that suddenly enters the field of view). However, unless it is later recognised as relevant for the set goal, thus becoming a GINP, it is discarded and becomes a non-GINP.

The difference between non-GINPs/pre-GINPs and temp-GINPs/GINPs accounts for the differences between subliminal/implicit/unconscious and supraliminal/explicit/conscious representations highlighted by many brain studies \citep{MeneguzzoTsakirisSchiothSteinBrooks2014}.
In addition, and importantly, these concepts can highlight the difference between \textit{awareness} and \textit{consciousness}. Indeed, non-GINPs and pre-GINPs can temporarily access consciousness, thus becoming temp-GINPs or stable GINPs that the agent can report about (awareness). On the other hand, only GINPs can be the target of active manipulation operations that, thereby supporting higher-order consciousness and flexible goal-directed behaviour.
Therefore, this distinction implies that awareness plays a `preparatory' role for conscious goal-directed processes, while consciousness involves the core operative stage of conscious goal-directed processes. In section `A GARIM agency scale' we refer to these different concepts as `phenomenal consciousness' (awareness), `access consciousness', and `manipulative consciousness'. The scale implies that Awareness and Consciousness are two poles of conscious goal-directed cognition.

\paragraph{GINPs and other definitions of conscious representations}

Many theories of consciousness expect that only specific representations have the properties needed to be consciously processed (see section `Comparisons of the GARIM theory with other theories' for more detailed comparisons). 
For example, the IIT proposes that conscious states show a high level of integration and specification. 
The GWT/GNWT propose that consciousness is supported by stable activations of highly integrated systems, dispatching information to short-range sub-modules. 
The Radical Plasticity Theory (part of the Higher-Order Theories of consciousness) suggests that only the meta-representations, featuring stability, strength and distinctiveness, are perceived as conscious contents. 
Finally, Neurorepresentationalism postulates that conscious states depend on multimodal/multi-level representations subserving goal-directed behaviours. 

While the GARIM theory focuses on a subset of conscious representations (i.e., those that support goal-directed behaviours), the definitions related to the different types of GINPs is compatible with those proposed by several theories of consciousness. In addition, however, the definition of GINPs explicitly requires their active goal-oriented manipulation to become stably conscious.

\subsection{Four key components at the basis of goal-directed manipulations of representations}

\begin{figure*}[htb!]
  \centering
   \includegraphics[width = 0.9\textwidth]{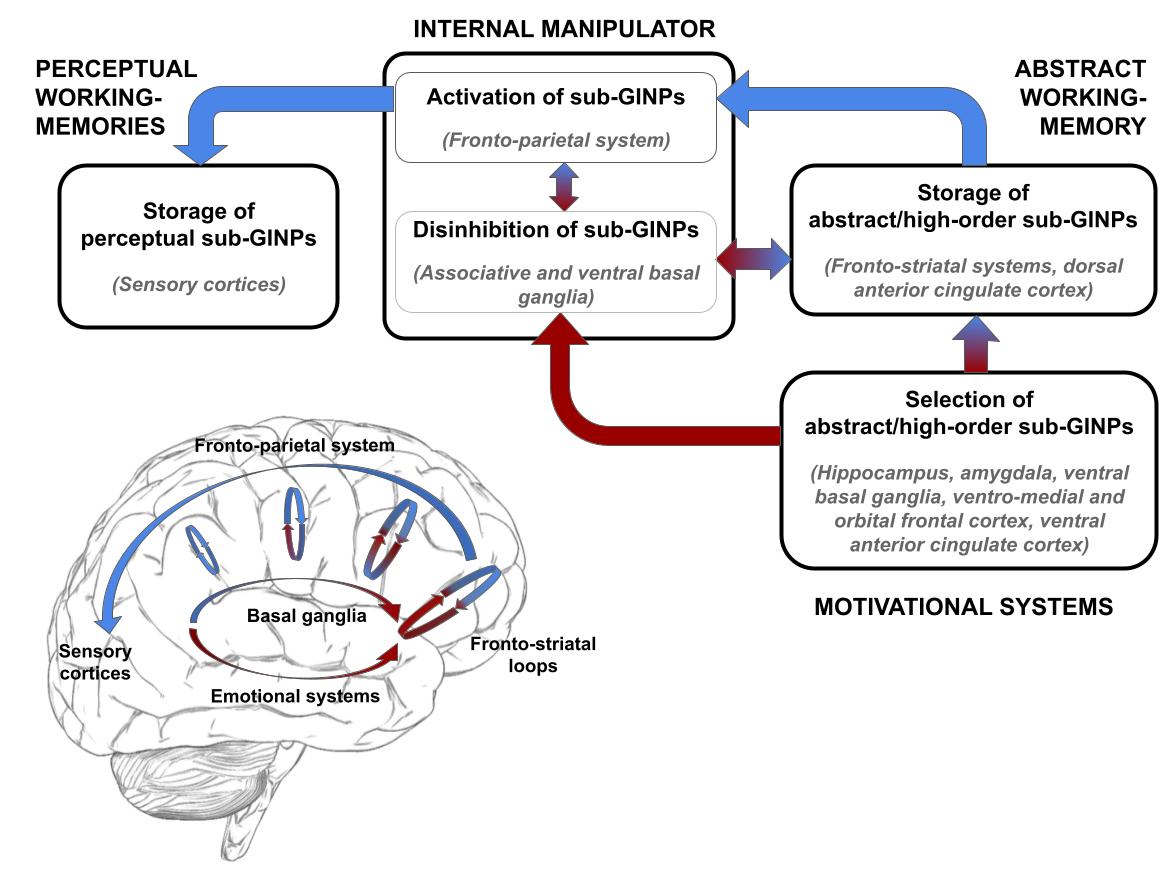}
     \vspace*{+5mm}
  \caption{
   Schema showing the `components' of the GARIM theory, and their relation with brain anatomo-functional systems.
   The red-to-blue coloured gradient indicates the decreasing involvement of motivational/emotional processes. 
}
  \label{Figure:4_components_model}
\end{figure*}

The GARIM theory postulates that conscious higher-order cognition is supported by goal-oriented manipulations of representations.
These manipulations rely on four key `components' (Figure \ref{Figure:4_components_model}), namely four partially overlapping anatomo-functional brain macro systems. Note that the GARIM theory inherits some key components from its precursor theory, extending them (in particular, by adding a fourth motivational component) and further specifying their functioning (e.g., see the section `The four classes of GARIM computational operations').

\paragraph{(1) Perceptual working memory component}

The three-component theory proposes that perceptual hierarchies play a key role in goal-directed behaviours. Moreover, various theories of consciousness postulate that hierarchical perceptual systems and perceptual working-memories support the emergence of conscious states (see section `Comparisons of the GARIM theory with other theories').

In the GARIM theory, the \textit{perceptual working memory component} plays a key role for the emergence of GINPs. The component corresponds to partially segregated `unimodal' sub-systems that perform bottom-up sensory processing. These operations support the formation of increasingly abstract perceptual sub-GINPs (e.g., from low-level features to high-level representations). In particular, the bottom-up information flows convey pre-GINPs/non-GINPs representations to higher-level cognitive areas. 
The component also supports a top-down information flow, causing the re-activation of the peripheral sub-GINPs (e.g., goal imagination and mental simulations during visual planning). These manipulation processes can contribute to transform pre-GINPs into GINPs and to inhibit temp-GINPs (non-GINPs that temporary access consciousness).
The component also implements peripheral modal working memories. These maintain active perceptual representations having a short duration and a high level of detail (e.g., the perceptual representation of a goal).


In the brain, the component is supported by cortical hierarchical pathways. These encode bottom-up information at multiple levels of abstraction, instantiating extensive associative networks linking sub-GINPs encoded in different cortices. 
At the same time, fronto-parietal cortical pathways activate in a top-down goal-directed fashion the contents of the modal working memories. 

\paragraph{(2) Abstract working memory component}

The three-component theory proposes that an abstract working memory plays a key role in goal-directed behaviours (e.g., storage of abstract goals). Moreover, most theories of consciousness ascribe a central role to working memory (see section `Comparisons of the GARIM theory with other theories').

In the GARIM theory, the \textit{abstract working-memory component} supports the active maintenance and integration of different goal-relevant sub-GINPs (e.g., related to contexts, behavioural strategies, predictions, and values). These sub-GINPs are related to low-level sub-GINPs (e.g., behavioural strategies can be related to movement representations and expected somatosensory and visual feedback) but encode more abstract information with respect to perceptual sub-GINPs. This feature makes them a form of meta-knowledge. 

Importantly, abstract sub-GINPs dynamically integrate both spatiotemporal relations of the world elements (e.g., own body parts, objects, other agents) and agent's predictions (e.g., goal-related action outcomes) in \textit{world models}. Thus, the abstract working-memory component exploits world models to monitor plans (i.e., check the prediction correctness based on percept and goals/sub-goals), to counter internal (self) and external (environment) disturbances, and finally to generate the missing knowledge (e.g., new points of view on objects, new solutions).

Within the brain, abstract multimodal sub-GINPs are encoded by different prefrontal cortices (e.g., dorsolateral PFC, ventrolateral PFC, and anterior cingulate cortex) and related subcortical areas (e.g., basal ganglia-thalamo-cortical loops and hippocampal system). Within each cortical area, neural winner-take-all mechanisms allow the activation of only one or few possible patterns at a time.
Importantly, the abstract working memory component plays a `hub role' by putting in relation sub-GINPs in different areas (e.g., different regions of the fronto-parietal network). 
In particular, it dynamically integrates abstract sub-GINPs with perceptual sub-GINPs, thus realising a close interaction with perceptual working memory. This coupling supports perceptual monitoring underlying conflict resolution, goal-alignment (e.g., perceptual monitoring of goal-prediction matching), and sub-GINP sequences activations at the basis of visual planning (images of world states traversed to reach the goal).

\paragraph{(3) Internal manipulator component}

The three-component theory proposes that goal-directed behaviours are supported by a top-down manipulator of representations. Several theories of consciousness attribute a central importance to attentional processes and their top-down influence on conscious information (see section `Comparisons of the GARIM theory with other theories'). 

The GARIM theory proposes that an \textit{internal manipulator component} manipulates the contents of abstract and perceptual working-memories. In particular, it selects and warps perceptual and abstract sub-GINPs to generate sequences of GINPs with increasing goal-alignment.  
Importantly, these manipulations support (a) monitoring and alignment of goals, sub-goals and world models (abstract sub-GINPs) with perceptions (perceptual sub-GINPs) to solve conflicts and internal/external disturbance and (b) the subsequent generation of new knowledge needed in case of novel situations and goals.

In the brain, the manipulator's operations are supported by two major selection mechanisms. One  corresponds to local inhibitory circuits of cortex, in particular those composing the cortical fronto-parietal system. The second corresponds to the disinhibition mechanisms of basal ganglia-thalamo-cortical loops. 
The influence of basal ganglia on the cortex has a diminishing gradient, moving from frontal to posterior cortical areas. Although the GARIM theory focuses on these manipulation brain systems, others could contribute to goal-directed manipulations. For example, the three-component theory proposes that the language system acts as an internal manipulator of abstract goals (e.g., inner speech; \citealp{granato2020computational, granato2022computational, granato2023flexible}).

\paragraph{(4) Motivational component} 

Motivational systems play a key role in the expression of goal-directed behaviours, from goal formation to action selection (see section `Beyond the three-component theory'). Various theories of consciousness take into consideration the role of motivational and emotional processes for consciousness (see section `Comparisons of the GARIM theory with other theories'). 

In particular, the GARIM theory proposes that a \textit{motivational component} indirectly guides the manipulator, contributing to select goals at different levels of abstraction within the abstract working memory.
Moreover, the motivational component also directly contribute to the manipulator operations, giving different salience to perceptual and abstract sub-GINPs (see Figure~\ref{Figure:4_components_model}).
To this purpose, the motivational component closely interacts with the perceptual and abstract working memories to perform goal-monitoring and goal-aligning operations based on the manipulation of percepts, world models and plans.

The motivational component also contributes to giving an emotional subjective nuance to conscious representations. In particular, perceptual sub-GINPs (e.g., representations of external stimuli and anticipated outcomes) are evaluated (appraisal) on the basis of their contribution to the achievement of goals (goal-alignment). This process contributes to integrate cognitive and emotional aspects of goal-directed behaviour, and plays a key role for the agent's subjective experience accompanying consciousness (see section `GARIM agency and the subjective experience of consciousness').

In the brain, motivational and emotional evaluations drive the selection processes of basal ganglia and cortical winner-take-all mechanisms. In particular, evaluations generated in subcortical structures (e.g., the hypothalamus, amygdala, hippocampus) reach the basal ganglia starting from the the limbic loop. Moreover, they reach various cortical areas starting from the PFC ventral areas (e.g., orbital, medial, and insular cortex).

\subsection{Four classes of GARIM operations}

\begin{figure}[htb!]
  \centering
     \includegraphics[width = 0.5\textwidth]{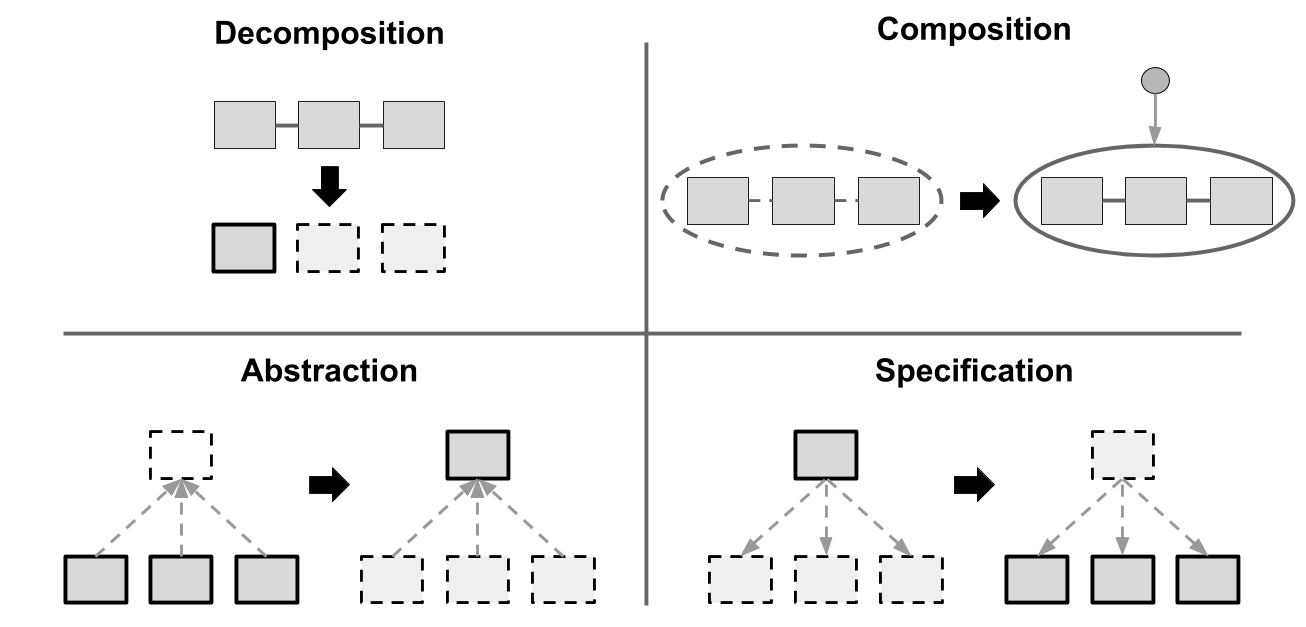}
     \vspace*{+5mm}
  \caption{The four classes of GARIM operations that the manipulator performs on GINPs.}
  \label{Figure:4_Primitive_img}
\end{figure}

The interaction of the four components supports goal-directed manipulations of internal representations.
These manipulations are divided in four \textit{GARIM operations} (Figure~\ref{Figure:4_Primitive_img}).
These operations modify the GINPs and are subjectively experienced by the agent as intentionally directed operations (see section `GARIM agency and the subjective experience of consciousness'). 
The four classes are now considered in detail.

\paragraph{(1) Abstraction}
Abstraction causes the generation of sub-GINPs at different levels of abstraction, from perceptual sub-GINPs to abstract sub-GINPs.
Abstraction also executes goal-dependent dimensional reductions, preserving only goal-relevant aspects of low-level sub-GINPs. 
For example, in addressing the goal `grasping the cup', abstraction operations might change the detailed sub-GINP related to the perceptual representation of the cup into a more abstract goal-oriented sub-GINP (e.g., a shape-based representation, ignoring colour because it is not useful for the pursued goal).

In the brain, abstraction relies on the hierarchically organised stages of cortical pathways.
Basal ganglia-thalamo-cortical macro loops (limbic, associative, motor) operate the selection of patterns at suitable levels of abstraction.

\paragraph{(2) Specification}
%
Specification performs the inverse operations with respect to abstraction. For example, starting from an abstract sub-GINP (e.g., `something to drink with') it can generate a sub-GINP corresponding to a specific object (e.g., `my preferred tea cup').

Since specification involves mappings from a few to many features, it requires a goal-directed and contextualised generation of suitable information (e.g., the perceptual details of `my preferred cup' when the goal is `drink tea at home'). These operations are made possible by the manipulator's selections and by the generative networks of perceptual and abstract working memories. 

In the brain, specification relies on the top-down `inverse' activation of cortical pathways, moving from multimodal representations in the frontal cortices to modality-specific representations in the lower sensory cortices. The generation of more detailed representations is guided by the cortical and basal-ganglia selection processes.

\paragraph{(3) Decomposition}
Decomposition performs the separation of representations (GINPs and sub-GINPs) into sub-parts. 
This operation executes a different kind of manipulation with respect to abstraction and specification. While the latter perform a `vertical manipulation' that changes the abstraction level, decomposition performs a `horizontal manipulation' at a fixed level of abstraction.
For example, decomposition could extract the representation of an object (e.g., `a tea cup') from the background, or the representation of a part of the object (e.g., `the handle') from other parts (e.g., `the cup container'). 

In the brain, decomposition could be supported by neural structures similar to those of specification, thus involving the cortex and basal ganglia-thalamo-cortical loops.
However, it might more strongly involve the channels and sub-channels within those loops to disinhibit specific cortical contents. Cortical local winner-take-all mechanisms should facilitate the selection of sub-parts of neural patterns. 

\paragraph{(4) Composition} 
%
Composition performs the inverse operations with respect to decomposition, integrating many sub-GINPs into larger sub-GINPs or into a coherent whole GINP.
Through composition, the agent can build global items starting from its parts (e.g., to consider a `cup container', `handle', `tea', and `tea spoon', as a whole `tea cup').

Composition supports various aspects of goal-directed processes. For example, it supports the generation of plans (e.g., by chunking a sequence of actions and their effects) or imaginary processes leading to solve a problem (e.g., building a new tool by aggregating various parts). 
Composition performs a different manipulation with respect to abstraction. Abstraction performs a dimensional reduction (loss of information) while composition `chunks representations' at the same level of abstraction.
However, composition and abstraction could give rise to adaptive synergies. For example, they could lead to integrating many sub-GINPs at the same abstraction level, then transforming the resulting sub-GINP into a more abstract one (e.g., chunking `reaching', `grasping', `transporting', `drinking' to generate the abstract goal `taking a tea').

In the brain, composition might rely on functional connectivity between different networks. 
Moreover, it might rely on physical connectivity linking semantically related neural patterns (e.g., two different colours within the visual cortex, or the `red' colour in the visual cortex and `alertness' in an affective area).

\paragraph{The integrated functioning of the GARIM operations: representation manipulations boost flexibility during problem solving} 

The GARIM operations give rise to a super-ordinate function we call \textit{Conscious Knowledge Transfer (CKT)}. CKT refers to a transfer of knowledge from familiar contexts to novel contexts, thus supporting flexible human cognition and behaviour. 
In particular, CKT operates by flexibly abstracting, specifying, decomposing, and composing the sub-GINPs that encode the current knowledge (e.g., related to objects, goals, actions, and expected outcomes).
Therefore, on the basis of multi-level goal-monitoring/goal-alignment evaluations, 
CKT allows the agent to generate the necessary knowledge to improve performance, to successfully act in changed conditions, or to accomplish novel goals.
Differently from the concept of \textit{generalisation}, CKT leads to the generation of new knowledge beyond previous experiences. While generalisation involves interpolation processes (e.g., the imagination of a goal position that involves an object positioned between two previously experienced positions), CKT involves extrapolation processes (e.g., the imagination of an object located anywhere in a known space; or the generation of a new tool based on composing elements). These operations are based on the extraction of relevant regularities from previous experiences, and their transformation to generate knowledge to address novel challenges (decisions, plans, problems).

Problem solving tasks are best suited to illustrate the CKT and the GARIM operations. 
Such problems are challenging because their solution requires the generation of missing knowledge on ill defined components.
For instance, consider the classic Duncker's problem \citep{Duncker1945OnProblemSolving}. In this task participants are required to fix a candle on a wall. They can only use some pins, available in a box, and some matches to solve the problem. The solution requires to pin the cardboard box on the wall and then set the candle on it.
This solution requires a `change of perspective' on the elements of the problem \citep{guilford1967nature, chrysikou2016functional}. Indeed, participants generally consider the box only as a container, but this change of perspective leads them to focus on its properties (e.g., `cardboard can be pinned'). Thus, they discover that the box can serve as a candle holder.
As highlighted by the `representational change theory' \citep{OhlssonInformationProcessingExplanationsofInsightandRelatedPhenomena}, the solution requires the participant to generate a new suitable representation of the key problem's sub-components (e.g., of the box).

The GARIM theory can explain the manipulation and generation of knowledge that leads to the solution of the Duncker's problem. For example, an agent could use decomposition to parse the scene, and then sequentially activate the sub-GINPs that encode the different objects of the task. 
When focusing on the cardboard, the agent might use decomposition and specification to analyse  the different feature-based sub-GINPs of the cardboard (e.g., the usual function, the shape, and the material). 
These sub-GINPs, potentially influenced by a context-dependent priming effect (e.g., a pre-GINP encoding the pin), can recall the representation of a previous experience (e.g., the agent that used pins to stick cardboard drawings on the wall). Exploiting composition, the agent might then transfer the piece of knowledge `cardboard things can be pinned on walls' (a sub-GINP) to the cardboard box (another sub-GINP). 
At last, the resulting sub-GINP could be abstracted (abstraction) and compared with the initial goal of `attaching the candle to the wall'. 
A high correspondence between the two would imply a high goal-alignment of the GINP, achieved thanks to the CKT.


\subsection{Subjective experiences during conscious goal-directed behaviours: the GARIM agency}

The nature of subjective experiences is widely debated in the literature, which commonly refers to them as the `hard problem of consciousness' \citep{Chalmers1995}. 
Although the GARIM theory does not offer a solution to the hard problem of consciousness, it proposes its own perspective on this topic. In particular, the theory relates the activity of the internal manipulator with the emergence of a subjective experience of agency. Therefore, we introduce the concept of \textit{GARIM agency} to identify the sense of agency that emerges during the expression of conscious flexible goal-directed behaviours. In particular, the theory proposes that the manipulation of representations generates an \textit{internal simulated reality} having three key features: self-models, 
emotional/perceptual vividness, and 
manipulation control.

First, the simulated reality involves some aspects of the agent itself.
This self-simulation can be enhanced based on previous experiences with other intentional agents \citep{fernandez2000executive}. 
%
Second, the manipulator activates low-level sub-GINPs that enrich the GINPs with detailed perceptual representations. The GINPs are continuously evaluated with respect to their goal-related alignment and thus they are emotionally charged. These GINPs hence exhibit perceptual and emotional features similar to those that the agent experiences when acting in the environment. For this reason, the internally simulated and manipulated reality is vividly perceived and felt   similarly to the real experience.
%
Third, the intentional manipulation of representations cause imagined effects similar to those caused by motor actions performed in the external environment.
Therefore, the manipulations produce a \textit{sense of agency} \citep{jeannerod2003mechanism} for which the agent perceives itself as the cause of `internal actions' (GARIM operations) and of the effects they produce. 

Note that the concept of GARIM agency is compatible with other concepts proposed by the literature. For example, \citet{metzinger2013myth} highlights the concept of mental action and cognitive agency to identify the capacity to control own goal-directed conscious processes. Moreover, mental action and self-control are concepts approached by the active inference framework \citep{Metzinger2017, hohwy2020predictive}. Thus, the GARIM theory captures many aspects of conscious goal-directed cognition and agency that are considered fundamental by other studies in the field.

\subsubsection{A GARIM agency scale}

\begin{figure}[htb!]
  \centering
     \includegraphics[width = 0.5\textwidth]{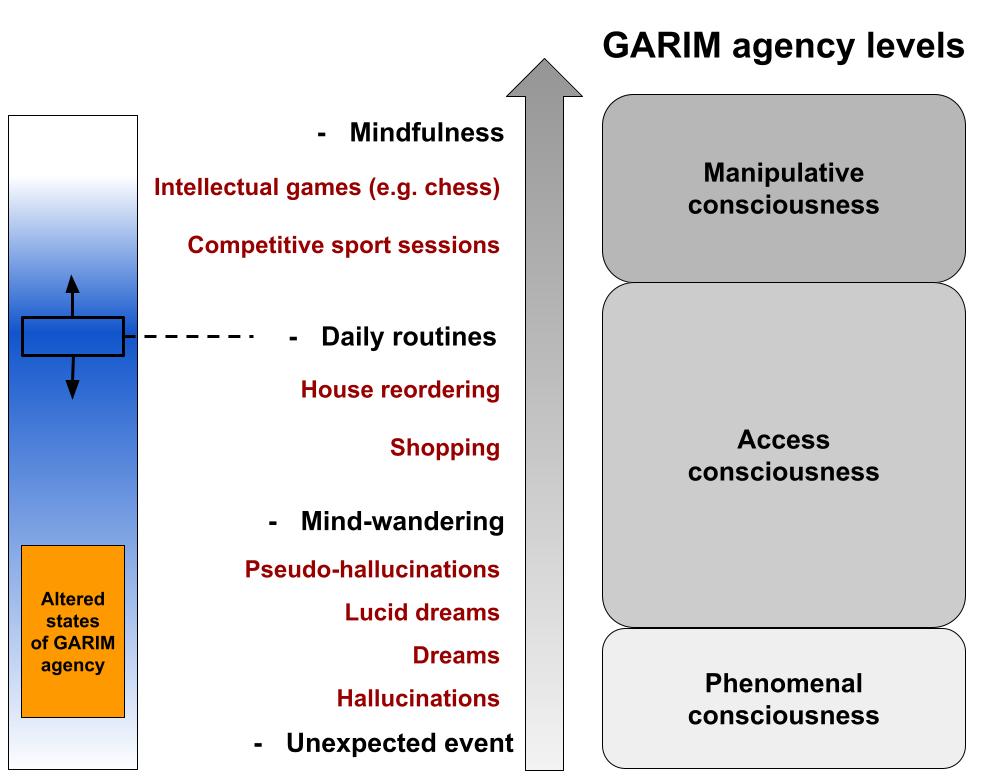}
     \vspace*{+5mm}
  \caption{A scale of consciousness based on the concept of \textit{GARIM agency}.
  }
  \label{Figure:SBJ_experience}
\end{figure}

The GARIM agency is a suitable concept for generating a quantitative scale, which takes into account the different levels of consciousness and flexible goal-directed behaviours. In particular, the three features of the GARIM agency (self-model, 
emotional/perceptual vividness, and 
manipulation control) lead to the emergence of three `levels of Consciousness' during the expression of goal-directed behaviours: 
phenomenal consciousness, 
access consciousness, and 
manipulation consciousness (Figure~\ref{Figure:SBJ_experience}). 
We explain these three levels by describing examples of human cognition. 

\textit{Phenomenal consciousness} pivots on the peripheral activations of perceptual/emotional sub-GINPs. They are triggered by either external perceptual inputs or internal bottom-up  processes (e.g., emotional/motivational events).
The emergence of an unexpected, goal-irrelevant perceptual event is an example of this GARIM agency level. 
Indeed, while possibly showing some emotional/perceptual vividness, this event activates a temp-GINP (conscious goal-irrelevant representation). It is accompanied by a low level of GARIM agency and it is soon discarded.
In case the representation is a pre-GINP (unconscious goal-relevant representation), it can be transformed into a GINP thus leading to a higher level of GARIM agency.

\textit{Access consciousness} involves a mild top-down selection that leads to a weak competition between different sub-GINPs.
This GARIM agency level can be exemplified by the state of \textit{mind-wandering} \citep{gruberger2011towards}.
This is a brain state, usually accompanying the performance of routines, that generates conscious sequential thoughts representing temp-GINPs (e.g., thoughts on possible actions).
In this respect, \cite{christoff2016mind} suggest that mind-wandering involves a shallow ``deliberate constraint'', that is, a partially deliberate cognitive control on own thoughts. 
Based on our proposal, this process should involve continuous transformations of pre-GINPs into GINPs and vice versa, and non-GINPs into temp-GINPs and vice versa.
These processes would be the effect of  a weak top-down control, and indeed mind-wandering can take place without awareness \citep{schooler2011meta}.

\textit{Manipulation consciousness} is characterised by a high control on internal representations.
This state is exemplified by specific forms of \textit{mindfulness} achieved in meditation
\citep{KabatZinn1990FullCatastropheLivingUsingtheWisdomofYourBodyandMindtoFaceStressPainandIllness,malinowski2013neural}.
For example, focused meditation aims to induce a high goal-directed attentional focus (e.g., on own breath).
This amplifies the access to consciousness of goal-relevant information (GINPs), and leads to a non-judmental state by strategically suppressing internal/external distractions (temp-GINPs) and ruminations
\citep{TangHoelzelPosner2015Theneuroscienceofmindfulnessmeditation,YatesImmergut2015TheMindIlluminatedACompleteMeditationGuideIntegratingBuddhistWisdomandBrainScience}. Note that even in the case of unfocused non-judgmental states (e.g., some forms of mindfulness) a higher-level goal can be active, namely `to keep the whole state of meditation intact from distraction'.
Similar features can be shared by brain states supporting a high attentional engagement in competitive sport sessions \citep{he2018sport, miller2011vision, memmert2009pay} or intellectual games (e.g., chess; \citealp{atherton2003functional, wang2020reduced, hanggi2014architecture}). 

Overall, the GARIM agency is expected to continuously fluctuate along the different levels of consciousness.
Healthy awake people might likely remain most of the time within middle levels of consciousness, for example when carrying out daily routines (e.g., house reordering and shopping).
The rest of the time they might have transitory phases into the lower levels of consciousness, and limited periods of time into the highest levels.
The following section proposes that there are states of consciousness accompanied by altered GARIM agency levels, falling between the middle and the low levels of consciousness.

\paragraph{Altered states of the GARIM agency}
\label{section:Altered_Sujective}

The GARIM theory and the scale presented in the previous section describe specific states of consciousness during the expression of goal-directed behaviours. Some of them may result in an altered state of GARIM agency (Figure~\ref{Figure:SBJ_experience}). 
For example, alterations of the GARIM agency could involve pseudo-hallucinations and hallucinations \citep{telles2015hallucinations}. 
Both states are experienced in the absence of external stimuli.
However, pseudo-hallucinations are perceived as unreal dummy perceptions whereas hallucinations are perceived as real perceptions.
Interestingly, the two show different levels of sensory controllability and vividness, which are higher in pseudo-hallucinations  \citep{van2001pseudohallucinations}. 
These evidence are compatible with an alteration of the GARIM agency.

Dreams and lucid dreams are other consciousness states that could involve an altered GARIM agency.
Dreams involve an uncontrolled imagination during the REM sleep while lucid dreams involve a partially controlled imagination \citep{stumbrys2012induction}. 
Both states correspond to the generation of a vivid internally simulated reality \citep{revonsuo2006inner}.
However, a higher level of control distinguishes lucid dreams from dreams \citep{voss2009lucid}, also suggesting that a stronger activation of frontal areas could cause this difference. 
Our proposal is compatible with this evidence as the alterations of the GARIM agency should depend on the influence of the top-down manipulator. 

\section{Comparisons of the GARIM theory with other theories}

The GARIM theory proposes an integrated framework that takes into account conscious and higher-order cognition, and thus it is compatible with most theories of consciousness \citep{seth2022theories}. 
In particular, it accounts for several key aspects that are considered fundamental for the emergence of conscious states (see Table~\ref{Table:Theories_comparison}).
Here we briefly describe these theories and we compare them with the GARIM theory. 

\paragraph{Integrated Information theory (IIT)}

The IIT \citep{Tononi2008Consciousnessasintegratedinformationaprovisionalmanifesto, TononiBolyMassiminiKoch2016Integratedinformationtheoryfromconsciousnesstoitsphysicalsubstrate, koch2016neural, Tononi2004AnInformationIntegrationTheoryofConsciousness} proposes that systems exhibiting high capacity of \textit{discrimination} (to encode several alternative neural representations of cognitive contents, e.g. percepts) and \textit{integration} (to encode several different associations between different aspects of neural representations, e.g.  stimuli) potentially have a high level of consciousness. 
This theory also proposes the $\Phi$ coefficient, a quantitative measure of the level of information integration. The thalamo-cortical system should have a key role in conscious states - thus an high  $\Phi$ - due to its high synaptic integration and interconnection. 
A recent update of the theory \citep{koch2016neural} has identified a `hot zone', located within the parietal cortex, that supports the formation of conscious contents. On the other hand, the frontoparietal system would have a control role of cognitive contents but not a central role for the emergence of a conscious state.

The GARIM theory does not delve into specific aspects of information theory, but it takes into account key features of the IIT such as discriminability and integration.
For example, the perceptual and abstract working memory components are expected to perform a high `discrimination' of experiences.
In particular, the manipulator component selects specific sub-GINPs between several alternative ones, thus assigning a specific and stable meaning to experiences (high discrimination).
At the same time, the generation of stable GINPs requires a dynamic highly flexible `assembling' of sub-GINPs based on suitable functional and anatomical connectivity (high integration).

The GARIM theory, however, has also important differences with respect to the IIT theory.
First, the IIT theory lacks a \textit{functional explanation} of conscious processes, fundamental for developing a comprehensive theory of consciousness \citep{Cerullo2015TheProblemwithPhiaCritiqueofIntegratedInformationTheory}. 
Indeed, computational systems can exhibit a high $\Phi$ while performing dull calculations \citep{SethIzhikevichReekeEdelman2006TheoriesandMeasuresofConsciousnessanExtendedFramework,aaronson2014not}.
Second, the GARIM theory emphasises the importance of a top-down and goal-directed manipulation of representations, while the IIT argues that a top-down control is not fundamental for the emergence of conscious contents. However, the two theories may focus on different conscious states. Indeed, the GARIM theory focuses on higher-order conscious states based on representation manipulation, but it expects the existence of conscious states with a temporary lower level of top-down control (temp-GINP). These latter states appear to be the focus of the IIT theory.

\paragraph{Convergence-Divergence Zones theory (CDZT)}

The CDZT \citep{Damasio1989Thebrainbindsentitiesandeventsbymultiregionalactivationfromconvergencezones, MeyerDamasio2009Convergenceanddivergenceinaneuralarchitectureforrecognitionandmemory,DamasioMeyer2009ConsciousnessanOverviewofthePhenomenonandofItsPossibleNeuralBasis} proposes that the brain is organised on multiple peripheral CDZs (P-CDZs; e.g.,  sensory cortices) and major central CDZs (C-CDZs; e.g, associative areas such as prefrontal, parietal, and temporal cortices). The P-CDZs transmit a bottom-up information flow to the C-CDZs, which perform a top-down retro-activation on them. 
In particular, the retro-activation increases the meaningful integration of bottom-up peripheral representations, resulting in conscious perception and imagination. Conversely, if the P-CDZs fail to activate the associated patterns in the C-CDZs, there is no retro-activation and the peripheral representations remain unconscious. 
The CDZs theory  also proposes that low-level somatic reactions assign an emotional valence to the representations within the C-CDZs, giving them sufficient priority to enter consciousness processing (`somatic marker hypothesis'; \citealp{bechara2005somatic, VerdejoGarciaPerezGarciaBechara2006Emotiondecisionmakingandsubstancedependenceasomaticmarkermodelofaddiction}). At last, the theory proposes that the representations at the basis of subjective experience (C-CDZs) encode sensorimotor relations between the agent, objects, and events in the external environment (`embodiment approach'; \citealp{DamasioMeyer2009ConsciousnessanOverviewofthePhenomenonandofItsPossibleNeuralBasis}).

The GARIM theory takes into account key elements of the CDZ theory \citep{Damasio1989Thebrainbindsentitiesandeventsbymultiregionalactivationfromconvergencezones}, further specifying them with neuroscientific and computational details.
The GARIM theory attributes a key role to the neural hierarchies of the brain. Indeed, P-CDZs and C-CDZs correspond to brain structures that should support perceptual and abstract sub-GINPs, respectively.
Furthermore, the GARIM theory proposes that these sub-GINPs are generated by bottom-up and top-down information flows.
Bottom-up flows support the encoding of perceptions in perceptual and abstract working memories at increasing levels of abstraction.
Top-down flows generate sub-GINPs that are functional to the achievement of goals. 
Both flows are controlled by the top-down manipulator, guided by motivations and goals. The manipulator selects the relevant information that travels along cortical hierarchies, thus improving the goal alignment of representations.

In line with the CDZT, the GARIM theory also takes into account the role of emotions and motivations for the assignment of valence to experience.
The GARIM theory specifies that motivational systems support sub-GINPs prioritisation through the GARIM computational operations.
Finally, in line with the CDZT, the GARIM theory emphasises the role of emotions and motivations for subjective experience.
In particular, it proposes that top-down manipulation processes activate peripheral sensory areas (imagination) and emotional/motivational systems (similar to the somatic marker hypothesis; \citealp{Damasio1989Thebrainbindsentitiesandeventsbymultiregionalactivationfromconvergencezones}).
The resulting activations then send a feedback to the central areas, associating a high level of perceptual vividness and emotional valence to subjective experience (see Section `GARIM agency and the subjective experience of consciousness'). 

\paragraph{Global Workspace Theory (GWT) and Global Neuronal Workspace Theory (GNWT)}

The GWT \citep{Baars1997, Baars2003, Baars2005, Baars2013} proposes that consciousness relies on a set of interacting cognitive elements, which are metaphorically associated with the elements of a theatre: conscious contents (e.g., percepts and thoughts; the `actors in the stage'); 
the \textit{global workspace} of working memory (the `theatre stage'); 
selective attention (the `theatre spotlight');
executive functions (the `director');
and unconscious background processes that interact with the global workspace (the `audience').
The theory proposes that alternative contents compete to enter the global workspace and thereby become conscious.
Selective attention processes, guided by the executive functions - in turn guided by motivations and goals - choose which conscious contents win the competition.
The winner contents are \textit{broadcasted} from the global space to the other processes to support higher-order processes (e.g., decision making and self-monitoring). 

The GNWT \citep{Dehaene1998, Dehaene2001, DehaeneChangeux2011Experimentalandtheoreticalapproachestoconsciousprocessing} was initially proposed to specify the neural correlates of the GWT based on extensive empirical support \citep{MashourRoelfsemaChangeuxDehaene2020} and computational formalisations \citep{DehaeneChangeux2005OngoingSpontaneousActivityControlsAccesstoConsciousnessaNeuronalModelforInattentionalBlindness, DehaeneLauKouider2017}.
The hypothesis proposes the existence of two computational spaces in the brain.
A first space is supported by high-density short/medium range connections and includes many specialised functional modules (e.g., sensory areas, motor systems, memory areas, evaluative components).
A second space, called `neuronal global workspace', is supported by long-range excitatory  projections \citep{DehaeneChangeux2011Experimentalandtheoreticalapproachestoconsciousprocessing} and includes a distributed set of associative areas (the prefrontal, parietal, temporal and cingulate cortices) and cortical-subcortical networks (e.g., the fibres of the corpus callosum and the cortico-thalamic system). 
This architecture allows the global workspace to generate global activation patterns with variable duration (\textit{ignitions}), involving distributed interconnected networks. These patterns strongly compete and inhibit or favour related patterns within peripheral specialised modules (e.g., perceptions, emotions and actions related to an object).
The frontal-parietal system plays a key role in supporting this top-down amplification of information. The biological underpinnings of the GNWT have been extended to envisage the existence of `buffers' (working memories) between the sensorial cortices and the neuronal workspace \citep{RaffoneSrinivasanvanLeeuwen2014Perceptualawarenessanditsneuralbasisbridgingexperimentalandtheoreticalparadigms,RaffoneSrinivasanvanLeeuwen2015RapidswitchingandcomplementaryevidenceaccumulationenableflexibilityofanallornoneglobalworkspaceforcontrolofattentionalandconsciousprocessingareplytoWybleetal}. At last, the GNWT has recently been integrated with inferential frameworks \citep{MashourRoelfsemaChangeuxDehaene2020}, suggesting that top-down amplification corresponds to an inferential process applied on bottom-up sensory inputs.

The GARIM theory integrates the main concepts of the GWT \citep{Baars1997} and the GNWT \citep{DehaeneChangeux2011Experimentalandtheoreticalapproachestoconsciousprocessing}.
In addition, it enriches those concepts by specifying the possible goal-directed computations (e.g., manipulation functions) and the brain mechanisms that might underlie them. 
First, as the GWT and GNWT, the GARIM theory assumes a `centre-periphery' architecture underlying conscious states as well as goal-directed behaviours.
In particular, it proposes multiple perceptual working memories that transmit information to the abstract working memory. Therefore, it integrates such information and dispatches the result back to the peripheral structures (the `broadcasting' of GWT and GNWT).
Second, the mechanisms underlying the generation of GINPs are compatible with those supporting `ignitions'.
Indeed, an ignition is a coherent activation of linked local neural patterns in central and peripheral areas.
Third, the GARIM theory ascribes a key role to the fronto-parietal brain system, proposing that it is fundamental for the top-down and goal-directed control of sensorimotor cortical pathways.

While sharing these important elements with the GWT and GNWT, the GARIM theory further specifies them.
First, in the GWT the patterns activated by ignitions are mainly generated by percepts. Instead, the GARIM theory postulates that the volitional goal-directed generation of GINPs depends on the selection of sub-GINPs by the top-down manipulator.
Second, while assigning an important role to the cortical fronto-parietal system, the GARIM theory highlights the pivotal role that the basal ganglia-thalamo-cortical system plays in the manipulation of sub-GINPs. 
Finally, the GARIM theory specifies the functioning of the bottom-up and top-down information flows in terms of computational manipulation operations (abstraction and specification/generative mechanisms).

\paragraph{Higher-order theories (HOTs)}

HOTs represent a family of theories originally formulated in philosophy (for a review, see \citealp{brown2019understanding}).
All HOTs share the idea that \textit{first-order representations}, for example the activation of patterns within the early stages of the visual cortex, are necessary but not sufficient to have a conscious experience. 
In particular, an agent can generate conscious contents only after first-order states have been evaluated and meta-represented by \textit{higher-order representations}.
The Radical Plasticity theory \citep{cleeremans2007consciousness, cleeremans2011radical}, an instance of the HOTs, proposes that meta-representations show three specific features: robustness, stability and distinctiveness. The theory has been recently integrated with inferential processes \citep{cleeremans2020learning}.
Most HOTs suggest that a certain level of `inner awareness' of one’s ongoing mental processes is necessary to have consciousness.
The claims of the HOTs have been supported by empirical evidence, highlighting the contribution of frontal networks in the formation of conscious higher-order representations \citep{lau2011empirical}. 
At last, the HOTs propose that first-order and second-order representations involve the interaction between subcortical and cortical systems, leading to an explanation of emotional aspects of conscious experience \citep{ledoux2017higher}.

The GARIM theory specifies the key concepts of the HOTs \citep{brown2019understanding} in terms of computational brain mechanisms.
First, the GARIM theory proposes that the interaction of four components leads to the encoding and selection of sub-GINPs at increasing levels of abstraction. 
The abstract sub-GINPs hence integrate the contents of lower-level perceptual sub-GINPs at a more abstract level, thus representing a form of meta-representations.
Moreover, in agreement with the Radical Plasticity Theory \citep{cleeremans2011radical}, GINPs should exhibit the three key features of robustness, stability and distinctiveness because they tend to (a) encode distinctive elements of goal-directed processes and (b) remain stable over time as long as they are relevant for the set goal. On the other hand, unconscious representations (e.g., non-GINPs) can briefly access consciousness (temp-GINPs) but then quickly fade away (low stability). 
Finally, the GARIM theory can also account for the `inner awareness' postulated by HOTs.
In particular, the goal-directed internal manipulation of representations give rise to a sense of agency that can accompany inner awareness (see Section `Subjective experiences during conscious goal-directed behaviours: the GARIM agency').

\paragraph{Sensori-motor theory (SMT)}

The SMT proposes that conscious experience pivots on the interactions between the brain, the body, and the environment  \citep{OReganNoee2001Asensorimotoraccountofvisionandvisualconsciousness, o2005sensory}.
The theory was developed within the theoretical frameworks of \textit{embodied cognition} \citep{Anderson2003, Garbarini2004, BorghiCimatti2010Embodiedcognitionandbeyondactingandsensingthebody} and \textit{enactivism} \citep{Hutto2005KnowingwhatRadicalversusconservativeenactivism}.
The theory substantially diverges from the other theories as it de-emphasises the role of brain processes and  representations, highlighting instead the importance of sensorimotor experience.
The theory proposes that \textit{sensorimotor contingencies}  (the events linking actions to sensory changes; \citealp{JacqueyBaldassarreSantucciORegan2019SensorimotorContingenciesAsAKeyDriveOfDevelopmentFromBabiesToRobots}) are fundamental in determining the phenomenal sensations that accompany conscious experience.
Differences in these sensorimotor activities distinguish sensory experience and reasoning/imagination processes.
In particular, sensory experience has `alertness' - the capacity to exogenously attract our attention - and `corporality' - the fact that bodily actions immediately modify the sensory input.


In agreement with the SMT \citep{OReganNoee2001Asensorimotoraccountofvisionandvisualconsciousness}, the GARIM theory supports the idea that consciousness plays a fundamental function for adaptation.
However, the SMT proposes that the key function of consciousness is the generation of a close coupling between motor action and its perceived effects. 
Instead, the GARIM theory proposes that the key function of consciousness is to enhance goal-directed processes to increase behavioural flexibility.
Moreover, the SMT pushes the embodied view of cognition towards anti-representationalist positions \citep{pennartz2018consciousness}.
The GARIM theory departs from these positions as `representations' and `manipulation of representations' are key concepts for it.
However, key theoretical aspects at the basis of both theories have been recently reconciled by highlighting that goal representations could support a link between actions and their perceived effects (sensory-motor contingencies) \citep{BaldassarreMannellaSantucciSomogyiJacqueyHamiltonORegan2018ActionOutcomeContingenciesAstheEngineofOpenEndedLearningComputationalModelsandDevelopmentalExperiments,JacqueyBaldassarreSantucciORegan2019SensorimotorContingenciesAsAKeyDriveOfDevelopmentFromBabiesToRobots, MannellaSantucciEszterJacqueyOReganBaldassarre2018Knowyourbodythroughintrinsicgoals}.
Finally, in line within the SMT, the GARIM theory clearly emphasises the importance of \textit{agency} for the generation of subjective conscious experience. However, the GARIM agency is centred on goal-directed representation manipulations while the SMT is focused on sensory-motor interaction with the environment.

\paragraph{Predictive Processing theories (PPTs)}

The PPTs (for a review, see \citealp{hohwy2020predictive}) are a family of theories that link conscious states to the concepts of `predictive coding', `error minimisation', and `world model'. According to the theory, the brain implements internal world models based on stacked dynamic neural loops. At each loop, the higher levels produce predictions about the activation of the lower levels (hence `predictive coding'), which in the lowest loops directly predict the percepts of the world. On the other hand, the lower levels flow information upward and compute prediction-errors by comparing the top-down predictions and their bottom-up activations. The prediction errors support a perpetual refinement of the world models. The Active Inference Framework (AIF), an instance of PPTs, proposes that the prediction-error can be minimised also by performing actions (e.g., ocular movements) to produce expected sensory data (predictive control). Overall, the brain correlates of PPTs correspond to the bottom-up hierarchical flows, for example from sensory cortices to prefrontal systems, and top-down feedback flows, from higher-order brain systems to lower sensory areas.

The GARIM theory shares some important elements with the PPTs. Both theories highlight the important role for consciousness of bidirectional brain hierarchies. 
In particular, the GARIM theory proposes that top-down information flows along the hierarchies implement generative processes reconstructing representations at lower-levels. This process is fully in line with the generative mechanisms of predictive coding. In addition, both the GARIM and the AIF explicitly refers to an active top-down control of action. 

Although these common points, the GARIM theory has additional elements and some divergent positions.
First, the GARIM top-down manipulator performs several goal-oriented operations on knowledge, at different abstraction levels. These processes generate new knowledge not only by interpolating previously acquired knowledge, but also by extrapolating it to produce more creative deviant representations (e.g., imagining a new tool). 
Instead, the PPTs and AIF, pivoting on the mechanism of prediction error, are more closely linked to interpolation processes.
Second, the GARIM theory ascribes a central role to goal-directed cognition and behaviour while initial PPTs proposals did not do so. However, recent proposals of the AIF have started to interpret goal-directed processes \citep{pezzulo2015active, matsumoto2022goal, hohwy2020predictive, friston2016active}, thus producing a potential common ground with the GARIM theory.
At last, although the GARIM theory ascribes a key role to information flows in cortex, it also proposes that basal-ganglia and cortical selection mechanisms play a key role to instantiate the manipulator's operations.

\paragraph{Neurorepresentationalism theory (NRT)}

The NRT \citep{Pennartz2015TheBrainsRepresentationalPoweronConsciousnessandtheIntegrationofModalities, pennartz2018consciousness, pennartz2022neurorepresentationalism} is a theoretical framework that defines Consciousness as a `multimodal and situational survey'. It proposes that conscious states depend on multimodal/multi-level representations, which are fundamental to sub-serve goal-directed behaviours (e.g., planning). This framework proposes five features that describe conscious experience: 
`multimodal richness', the emergence of sensations in multiple distinct modalities; 
`situatedness/immersion', the sensation that our body is immersed in the space and has a central position with respect to the surrounding stimuli;
`unity/integration', the emergence of a single undivided and multi-modal representation;
`dynamics/stability', the emergence of dynamic perceptions (e.g., external environment changes) and static perceptions (e.g., stationary objects);
`intentionality', the generation of signals that are interpreted as something other than ourselves. At last, the NRT highlights that predictive processing is a suitable framework for describing the neuro-computational basis of conscious states. In particular, it proposes that multi-level representations emergently lead to multimodal and spatially wide
`super-inferences', corresponding to phenomenal experiences.

The GARIM theory and the NRT share the idea that conscious states have the scope to support goal-directed behaviours. Notably, they are the only two theories that explicitly and systematically propose this bridge. Moreover, the two theories share some hallmarks that describe conscious states. In particular, some features of conscious representations proposed by the NRT are consistent with the definition of GINPs. For example,  GINPs are defined as integrated systemic representations whose sub-parts encode different aspects of goal-directed processes (e.g., motivations, perceptions, actions). Moreover, GINPs are formed by sub-GINPs at different levels of abstraction, from modality-specific working memories to multi-modal abstract working memories. At last, GINPs are `embodied representations' that integrate aspects of the environment, the agent and their relationships (e.g., action outcomes). Therefore, the `multimodal/multilevel representations' proposed by the NRT partially overlaps with the concept of GINPs.  

Despite these commonalities, the GARIM theory shows key differences with the NRT. First, GINPs and sub-GINPs are characterised by a specific dimension defined `goal-relatedness' (i.e. their relatedness with respect to the set goal), not considered by the NRT.
Second, the two theories tend to focus on two different aspects of goal-directed behaviour. In particular, the NRT focuses on the emergence of the best representations which then subserve goal-directed processes. Instead, the GARIM theory mostly focuses on the representation manipulation operations that constitute goal-directed processes and behaviour. 

\subsection{Integrating key aspects of Consciousness into a neuro-functional framework of flexible goal-directed behaviour.}

\begin{table*}[hbt!]
\centering
\resizebox{\textwidth}{!}{
\begin{tabular}{|c | c | c | c | c | c | c | c | c|}
    \hline
{\textbf{\thead{\\\\\\\\ \Large{Theories}}}} & \multicolumn{8}{c|}{\textbf{\Large{Key concepts}}} \\
 \cline{2-9}
      & \textbf{\thead{Information \\ integration and \\ discrimination}} & \textbf{\thead{Hierarchical \\ bidirectional \\ flows}} & \textbf{\thead{Broadcasting, \\ Ignitions}} & \textbf{\thead{First/second order\\ representations, \\ Inner awareness}} & \textbf{\thead{Embodiment, \\ Sensorimotor \\ contingencies}} & \textbf{\thead{Predictive \\ inferential \\ processes}} &
      \textbf{\thead{Multi-modal and \\ multi-level situated \\ representations}} &
      \textbf{\thead{Goal-aligning \\ representation \\ manipulations}}\\
    \hline
    \thead{IIT}
    & \ding{51} \ding{51} & \ding{55} & \ding{55} & \ding{55} & \ding{55} & \ding{55} & \ding{55} & \ding{55} \\
    \hline
    \thead{CDZT} %
    & \ding{55}/\ding{51} & \ding{51} \ding{51} & \ding{55} & \ding{55} & \ding{51} & \ding{55} & \ding{51} & \ding{55} \\
    \hline
     \thead{GWT/GNWT}   
     & \ding{55} & \ding{51} & \ding{51} \ding{51} & \ding{55} & \ding{55} & \ding{51} & \ding{55}/\ding{51} & \ding{55}/\ding{51} \\
    \hline
     \thead{HOTs} 
     & \ding{55} & \ding{55}/\ding{51} & \ding{55} & \ding{51} \ding{51} & \ding{55} & \ding{55}/\ding{51} & \ding{55} & \ding{55} \\
    \hline
     \thead{SMT} 
     & \ding{55} & \ding{55} & \ding{55} & \ding{55} & \ding{51} \ding{51} & \ding{55} & \ding{55} & \ding{55} \\
    \hline    
    \thead{PPTs} & \ding{55}/\ding{51} & \ding{51} & \ding{55} & \ding{55} & \ding{55}/\ding{51} & \ding{51} \ding{51} & \ding{55}/\ding{51}  & \ding{55} \\
    \hline   
    \thead{NRT} & \ding{55}/\ding{51} & \ding{51} & \ding{55} & \ding{55} & \ding{55}/\ding{51} & \ding{51} & \ding{51} \ding{51} &  \ding{55} \\
    \hline  
    \textbf{\thead{GARIM theory}} & \ding{51} & \ding{51} & \ding{51} & \ding{51} & \ding{51} & \ding{51} & \ding{51} & \ding{51} \ding{51}  \\
    \hline
\end{tabular}
 }

\vspace{5mm}
    \caption{Main concepts of the theories of consciousness considered in this work. 
    Symbols:
    \ding{51} \ding{51}: 
      concept pivotal for this theory; 
    \ding{51}: 
      concept compatible/encompassed by this theory; 
    \ding{55}/\ding{51}: 
      concept partially compatible/encompassed by this theory;
    \ding{55}: 
      concept not compatible/encompassed by this theory.}
    \label{Table:Theories_comparison}
    
\end{table*}

Recent works propose an analysis and comparison of the main theories of consciousness \citep{seth2022theories, del2021comparing}. Above we have compared these theories with the GARIM theory, highlighting their similarities and differences (for a summary, see Table~\ref{Table:Theories_comparison}). 
Due to its functionalist systemic approach, the GARIM theory contributes to integrate the other theories on consciousness at two levels: a `background integration' level and a `focused integration' level.
Concerning the first level, multiple concepts from previous theories describe `brain functioning/organisation principles' with which the GARIM theory is compatible (e.g., `information integration/discrimination', `embodiment and sensory-motor contingencies', and `predictive inferential processes').
Concerning the second level, some concepts proposed by other theories are central also for the GARIM theory (e.g., `first/second order representations and inner awareness', `hierarchical bidirectional flows', `broadcasting/ignitions', and `multi-modal/multi-level situated representations'). 

Overall, the GARIM theory highlights that higher-order cognition and consciousness necessarily require all these elements. However, the two different levels of integration could differentially benefit scientific and technological fields. For example, elements of the focused integration could aid the design of computational models. Indeed, they usually reproduce specific functions and neural mechanisms to explain brain and behaviour. On the other hand, elements drawn from background integration could aid the design of AI/robotic architectures. Indeed, these systems can also benefit from algorithmic solutions that are only conceptually inspired by higher-order human cognition and consciousness (see section `Implications of the GARIM theory for computational modelling, AI and Robotics').


\section{Experimental and clinical implications of the GARIM}

The GARIM theory represents a theoretical framework that has implications for several fields. 
In this section we first consider its contribution to the understanding of the concept of `Intelligence'. Then we proposes interpretations of psychological and neuropsychological evidence on goal-directed behaviours and consciousness.

\subsection{GARIM theory and Intelligence}
\label{section:RIM_Intel}

The GARIM theory focuses on the higher-order goal-directed cognition involving conscious states. These same processes could be at the basis of the expression of `intelligent processes and behaviours'.
Although the investigation of intelligence is beyond the scope of this work, the GARIM theory can contribute to its understanding, in particular to clarify its relationship with flexible goal-directed cognition and consciousness.

The term `intelligence' refers to a composite construct encompassing multiple areas of competence \citep{gardner2000intelligence} and is measured with different scales of intelligence (e.g., WAIS; \citealp{benson2010independent}). 
Recently, new theoretical frameworks have stressed the difference between domain-general and domain-specific intelligence \citep{burkart2017evolution}, also strengthening the relationship between intelligence and goal-directed behaviour \citep{chiappe2005evolution, tegmark2017life}.

In our previous computational proposals we modelled the interaction between domain-general processes (e.g., working memory and motivational systems) and domain-specific competence (e.g., sensory and motor learning).
This allowed the study of task-related representation learning \citep{granato2022Integrating} and goal-directed representation manipulation \citep{granato2021internal, granato2020computational, granato2022computational, granato2023flexible}. 
On the basis of these works, we explicitly proposed the idea that the flexibility characterising domain-general intelligence rests on the goal-directed manipulation of representations \citep{baldassarre2020goal}.
The GARIM theory extends these ideas to higher-order cognition and consciousness.
In particular, it proposes that consciousness boosts flexibility, a key aspect of general-domain intelligence.
This flexibility might aid the acquisition of domain-specific competences (e.g., motor skills) through the top-down guidance of the learning processes. Furthermore, flexibility might support the on-shot selective performance of previously acquired automatic behaviours.

This proposal is compatible with some important features of other theories of consciousness.
For example, the Radical Plasticity Theory (belonging to the HOTs) suggests that consciousness boosts learning processes. 
Moreover, the GWT and the GNWT suggest that the global-workspace information broadcasting improves the local learning of representations within peripheral brain sub-modules (e.g., motor modules). 
Moreover, the proposal is compatible with the concept of information integration proposed by the IIT. Indeed, flexible intelligent behaviour should require a high information integration within higher-order brain areas (e.g., the abstract working-memory), in turn influencing the lower-order ones (e.g., the motor and perceptual areas). 


\subsection{An interpretation of experimental and clinical evidence based on the GARIM theory}

\label{sec:RIM_Experiments}

The GARIM theory may be useful in interpreting psychological and neuropsychological evidence on goal-directed behaviours and consciousness.
Furthermore, it may stimulate the development of new experimental paradigms investigating the functional role of conscious states in flexible cognition.

\subsubsection{Lesion studies, goal-directed cognition, and conscious states}

The relationship between brain lesions and consciousness disorders is still not fully clear.
In particular, there is no research that systematically links impairments of frontal systems and basal ganglia, which play a key role for our proposal, with consciousness disorders.
However, the empirical support of HOTs indicates that PFC lesions cause a deficit in consciousness-related processes (e.g., metacognitive capabilities; \citealp{lau2011empirical}).
Moreover, recent proposals suggest that PFC lesions could influence consciousness in unnoticeable ways \citep{fox2020intrinsic}. 
On the other hand, various studies show that basal ganglia lesions cause a general consciousness impairment (e.g., \citealp{rohaut2019deep}).
Moreover, a bulk of studies \citep{ell2006focal, ell2010rule, ward2013functional, price2009rule} show that focal damages of basal ganglia impair explicit/conscious reasoning but not implicit/unconscious categorisation. 
These studies do not explicitly investigate consciousness, but they put in relation impairments of key elements of goal-directed cognition and explicit/conscious processes.

In general, the GARIM theory does not propose a conclusive explanation regarding the relation between consciousness disorders and frontal/basal ganglia lesions.
However, it proposes a link between these lesions, the possible alterations of explicit/conscious cognitive processes and goal-directed flexible behaviours.
For example, the GARIM theory predicts that extended lesions to PFC systems and associative portions of basal ganglia would impair abstract working memory and the top-down manipulator. Their impairment should corrupt the manipulation of GINPs. In particular, this alteration should lead to a reduced ability to transform pre-GINPs in GINPs or to suppress temp-GINPs (e.g., distractors). Therefore, these alterations should make an agent less generative and focused. These predictions are consistent with the clinical literature on goal-directed behaviour. Indeed, alterations of PFC and basal ganglia cause cognitive inertia, namely a reduced capacity to intentionally generate/activate strategies required to successfully complete a given program of actions \citep{levy2012apathy, levy2021pathology}. Despite these studies do not explicitly refer to consciousness, our theory proposes a link between conscious states and the above deficits. Indeed, cognitive inertia should alter the generation of GINPs, thus the exploitation of conscious and generative processes to express flexible behaviours. Moreover, the impairment of these structures could alter the GARIM agency, explaining the emergence of hallucinatory perceptual representations after frontal and basal ganglia lesions \citep{fornazzari1992violent, frith1996role, wodarz1995musical, mcmurtray2014acute}.

Note that these alterations do not correspond to global alterations of consciousness (vigilance/awareness). Indeed, the GARIM theory predicts that a focused lesion of PFC systems and basal ganglia would not cause a general loss of consciousness (e.g., coma).
Moreover, they would not prevent the access of stimuli to consciousness (phenomenal consciousness). 

In summary, the GARIM theory predicts that the frontal cortex and basal-ganglia impairments alter the link between consciousness and flexible behaviours. In particular, they impede an adequate emergence/management/manipulation of GINPs (access consciousness and manipulative consciousness). This corresponds to an inefficient exploitation of conscious processes for generating new knowledge and new perspectives during the expression of goal-directed behaviours (e.g. problem solving).

\subsubsection{Experimental evidence: the predictions of the GARIM theory}

The GARIM theory does not yet have direct empirical support, but it produces specific experimental predictions. Importantly, these predictions are in line with the experimental evidence provided by other theories of consciousness.

First, the GARIM theory predicts that perceptual sub-GINPs, involving the posterior higher-order sensory cortices, should remain active throughout the performance of explicit tasks.
These activations should support bottom-up abstraction and top-down generative processes.
This prediction matches the experimental evidence at the basis of the IIT.
Indeed, by contrasting stimulation effects during coma and wakefulness, evidence shows that a sustained activation of the posterior `hot-zone' is necessary for consciousness \citep{koch2016neural}.

Second, the GARIM theory also predicts that the emergence of GINPs is preceded by the activity of the top-down manipulator, involving the synergistic activation of the fronto-parietal system and the basal ganglia.
This prediction agrees with the evidence produced by the GNWT on contrastive tasks (e.g., masking, binocular rivalry, attentional blinking; \citealp{aru2012distilling}), highlighting that consciousness emerges due to a strong activation of the fronto-parietal areas (`ignitions'; \citealp{Dehaene2011}).
The activation of the top-down manipulator and the emergence of GINPs would correspond to the ignition processes recorded in these studies.
In addition, the GARIM theory further predicts that, given the same stimuli, different ignitions (GINPs) would emerge when different goals are pursued.

Third, some studies argue that there can be a dissociation between attention and explicit/conscious processing \citep{koch2007attention}.
These proposals are usually linked to bottom-up attention rather than top-down attention.
Indeed, attention processes are generally considered `necessary' to pass from unconscious to conscious processing \citep{van2010consciousness, raffone2014interplay,pitts2018relationship}, but they may not be `sufficient'.
In this respect, the GARIM theory predicts that: (a) stimuli having a high relevance for the pursued goals have a higher chance to be selected by attention and thus to access consciousness (pre-GINPs); (b) stimuli with a high bottom-up saliency may be able to enter consciousness (temp-GINPs) but they fade in case of a lack of support from top-down goal-directed mechanisms.

Finally, the GARIM theory predicts that a basal-ganglia/prefrontal cortex activation is necessary to generate a goal representation.
This prefrontal activation precedes and guides the GINP generation and conscious goal-directed behaviour.
This prediction agrees with evidence reported by the HOTs.
In particular, these show that a prefrontal activation is necessary to support second-order activations and the evaluation of own knowledge \citep{lau2011empirical}.
Our proposal agrees with these interpretations, as GINPs involve second-order representations integrating perceptual, motivational, and motor representations.
Moreover, the GARIM theory specifies that conscious processes involve both the manipulation of representations and the evaluation of their alignment with the pursued goal.

Overall, however, we believe that the tasks on consciousness proposed so far can only partially test the basic principles of the GARIM theory. The next section elaborates on this idea.

\subsubsection{Towards new tasks and protocols that test GARIM theory more directly}

Notwithstanding the growing evidence, empirical support of the major theories of consciousness is still unsatisfying \citep{yaron2022contrast, del2021comparing, doerig2021hard, melloni2021making}.
The GARIM theory can contribute to identify the problems that prevent the collection of more solid empirical evidence on consciousness.

Common experimental protocols (e.g., contrastive methods; \citealp{aru2012distilling}) mostly focus on the first stage of conscious processing considered by the GARIM theory, requiring experimental participants to \textit{detect a stimulus} and to perform simple actions in response to it (e.g., reply `yes/no' or choose one between few options, e.g. by voice or by pressing buttons).
According to the GARIM theory, these tasks are not sufficient to test manipulative consciousness. 
In particular they focus only on \textit{awareness}, explained by the GARIM theory as the initial passage from non-GINPs/pre-GINPs to temp-GINPs (phenomenal consciousness).
Instead, these tests are not sufficient to dissociate phenomenal consciousness and manipulative consciousness, the latter of which requires sustained manipulative processes.
Indeed, experiments capable of making this distinction should involve new goals or new conditions that require goal-directed processes (planning or problem-solving).
Alternatively, they should require the re-evaluation of relationships directed to increase goal-alignment, for example in relation to action-subgoal or subgoal-goal relationships.
For example, \cite{Weiskrantz1995TheProblemofAnimalConsciousnessinRelationtoNeuropsychology} discussed a possible experimental approach potentially testing goal-directed conscious processes. The author considers how blindsight patients can successfully discriminate stimuli without awareness \citep{poppel1973residual, weiskrantz2004roots}. Moreover, paraplegic patients can produce limb responses again without awareness \citep{weiskrantz1991disconnected}. With both these patients, `commentary actions' (e.g., `press a button' or `verbally report your experience') are necessary to check the presence of awareness.  
Similarly, to test these processes in animals it is necessary to pre-train them in the use of commentary actions (e.g., press a button; \citealp{cowey1995blindsight}).
The key point is that both humans' and animals' commentary actions might involve habitual processes rather than intentional conscious processes.
To avoid this problem, new experimental paradigms have been developed to explicitly test the presence of goal-directed processes (e.g., devaluation; \citealp{BalleineDickinson1998Goaldirectedinstrumentalactioncontingencyandincentivelearningandtheircorticalsubstrates,MannellaMirolliBaldassarre2016GoalDirectedBehaviorandInstrumentalDevaluationANeuralSystemLevelComputationalModel}). This proposal supports the idea that an effective experimental verification of goal-directed manipulations of representations can be a key step to check the operation of consciousness. 

We started to investigate the concept of representation manipulation with computational models  \citep{granato2020computational,granato2021internal, granato2022computational, granato2023flexible} by using the Wisconsin Card Sorting test \citep{heaton1993wcst}.
Even if this test measures executive functions and not consciousness, it involves an explicit categorisation and requires important representation manipulation processes (e.g., the selection of different representations to best support a flexible goal-direct behaviour in a changing environment).
Despite its relevant features, however, the test is not yet able to check various aspects of consciousness considered relevant by the GARIM theory (e.g., multi-stage planning or problem solving).

Overall, adequate tasks testing the GARIM processes should complement existing paradigms focused on testing awareness. 
In particular, an ideal task should have these elements:
(a) test perceptual \textit{awareness}, for example require to identify/categorise input patterns based on explicit rules; 
(b) request the achievement of \textit{new goals}, face \textit{new conditions}, or \textit{improve goal-alignment} so as to require the internal manipulation of representations to \textit{produce new knowledge};
(c) test the specific use of the \textit{GARIM manipulation operations} (abstraction, specification, decomposition, composition);
(d) test the processes of \textit{monitoring of goal-alignment};
(e) test the key elements of the GARIM agency: \textit{self-model, 
emotional/perceptual vividness, and manipulation control}.

\section{Implications of the GARIM theory for computational modelling, and for AI and Robotics}

The GARIM theory takes into account both neural and computational aspects of conscious and goal-directed behaviour. 
Indeed, it has both scientific and technological implications. 
%
First, the theory paves the way to the development of new computational models. In particular, these could capture computational operations at the basis of conscious and flexible goal-directed behaviour (e.g., top-down manipulation) and related neural mechanisms (e.g., competitive cortical and sub-cortical selection mechanisms).
These models could produce quantitative predictions to be tested against specific empirical data, thus corroborating our theory.
Second, the theory provides a guideline to possibly enhance the current AI and robotic systems. These systems might be empowered with functions and components proposed by the GARIM theory. The upgraded systems should be evaluated for their ability to improve performance with respect to current systems (e.g., in terms of goal-oriented flexibility and learning speed).


\subsection{Towards computational models of the GARIM theory}
\label{sec:computationalmodels}

\begin{figure}[htb!]
  \centering
     \includegraphics[width = 0.5\textwidth]{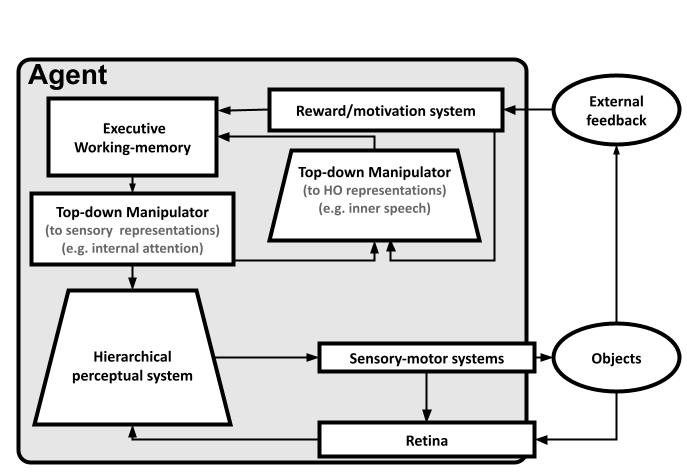}
  \caption{Schema of an already published computational model of the three-component theory \citep{granato2021internal, granato2020computational, granato2022computational, granato2023flexible}. The model is a starting point for building GARIM-inspired computational models.
    }
  \label{Figure:Comp_arch}
\end{figure}

We already operationalised the three-component theory with a computational model (Figure~\ref{Figure:Comp_arch}; \citealp{granato2021internal, granato2020computational, granato2022computational, granato2023flexible}). This model is supported by a neuro-inspired system architecture based on machine learning elements (generative models, recurrent neural networks, and reinforcement learning) and novel brain-inspired algorithms. The model was validated with human experimental data in various conditions (e.g., frontal patients, Parkinson, Autism) and ages (e.g., children, teenagers, young adults, and middle adults).
In particular, the model reproduced data from many cohorts of human participants performing the Wisconsin card sorting test (WCST; \citealp{berg1948simple,heaton1993wcst}). 
We used the WCST to test the model because, although it was initially proposed to test executive functions in general, it has now become the most commonly used neuropsychological test of cognitive flexibility \citep{miles2021considerations}. 
In this respect, the test requires a top-down switching of internal representations to successfully accomplish a goal when the environment changes \citep{granato2021internal}. 

\begin{figure*}[htb!]
  \centering
     \includegraphics[width = 0.9\textwidth
     , height = 27 EM
    ]{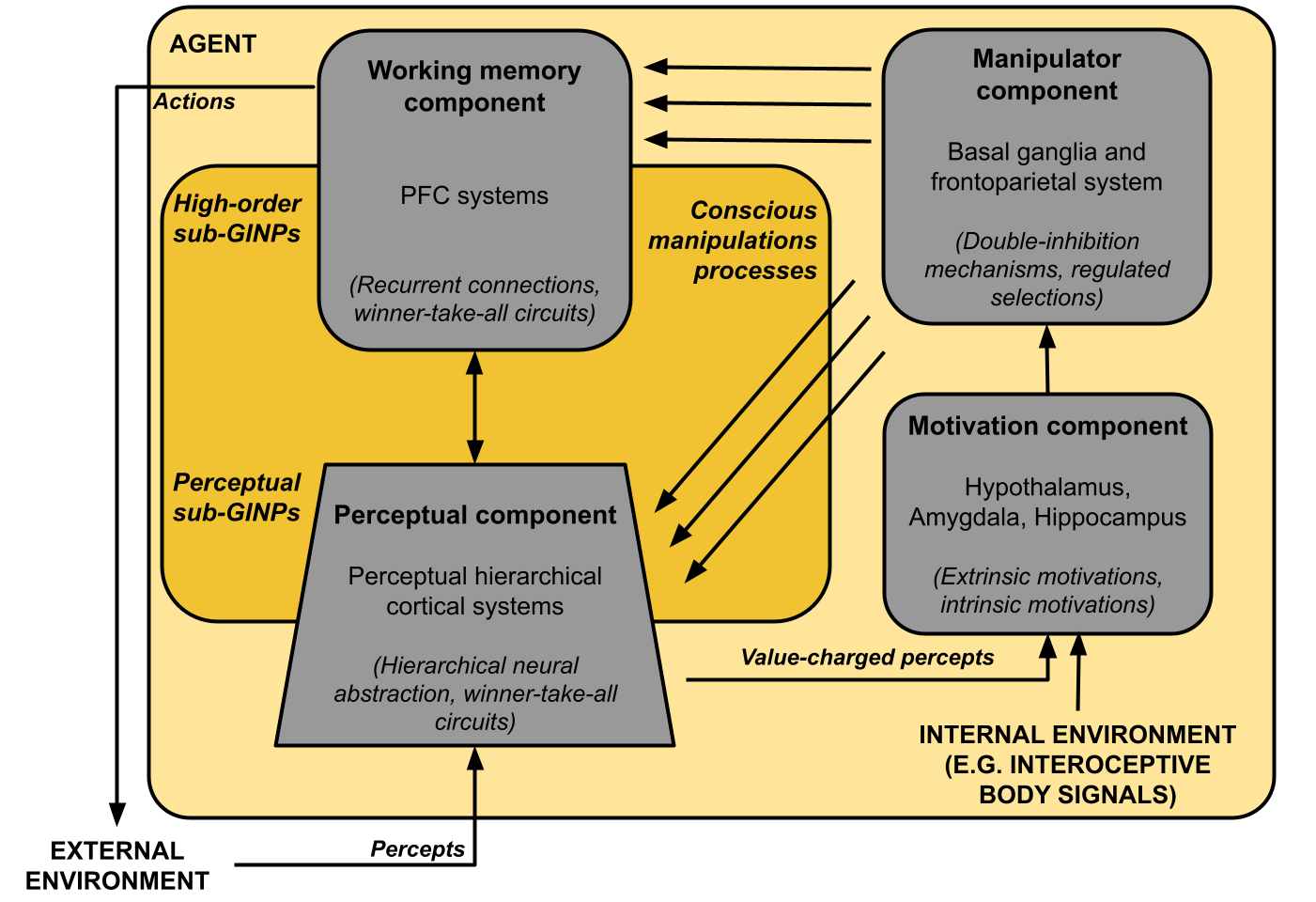}
     \vspace*{5mm}
  \caption{The GARIM blueprint architecture aiding the realisation of specific implementations of models following the GARIM theory principles. Italics in brackets: main brain neural mechanisms (structure and processes) possibly implementing the components.
  }  \label{Figure:RIMBlueprintArchitecture}
\end{figure*}

Overall, the model emulates human flexible goal-directed cognition and behaviour.
Since it is based on the three-component theory, it emulates three of the four components postulated by the GARIM theory (a hierarchical perceptual system, an executive working memory, and a top-down manipulator), sensory-motor loops, and first-order/second order representations and manipulations. 
Although the model specifically aimed to solve the WCST and did not consider conscious processing, it could still capture the C1 and C2 levels of simulation proposed by machine consciousness \citep{gamez2008progress}. In particular, the model shows an explicit rule-based categorisation process relevant for consciousness functions. Moreover, the model presents various architectural and functional elements supporting consciousness in the brain.
For these reasons, this model represents a possible starting point for building new computational models following the principles of the GARIM theory.
To this end, we now propose a `blueprint architecture' giving guidance to this purpose (Figure~\ref{Figure:RIMBlueprintArchitecture}).

The architecture should be supported by an adequate interaction of the four key components (perceptual hierarchies, working memory, manipulator, and motivation components).
However, their implementation could follow different approaches that emulate the brain mechanisms at different levels of detail.
For example, the models could be implemented with neuro-inspired algorithms (e.g., neural networks and reinforcement learning methods) abstracting from the details of the brain mechanisms (as we done in \citealp{granato2021internal}).
On the other hand, the models could emulate finer biological details of the brain (e.g., spiking neurons and neuronal connectivity; \citealp{DehaeneChangeux2005OngoingSpontaneousActivityControlsAccesstoConsciousnessaNeuronalModelforInattentionalBlindness,DayanAbbott2001TheoreticalNeuroscience}). The following paragraphs examine potential approaches to implement each component of blueprint architecture.

\subsubsection{Perceptual component: generating perceptual and higher-order GINPs}

The perceptual hierarchical component should be able to perform both abstraction and specification based on generative mechanisms.
Deep Belief Networks (DBNs; \citealp{hinton2006fast,Hinton2012}) are suitable for implementing this function.
They can learn input representations at increasing levels of abstraction based on statistical regularities and task demands \citep{granato2022Integrating}.
Moreover, they are able to generate representations on the basis of previous inputs and top-down generative processes \citep{granato2021internal}.
\textit{Spiking-neuron neural networks} are another approach that can be used to perform representation learning of key elements and timed chains
\citep{KappelNesslerMaass2014STDPinstallsinWinnerTakeAllcircuitsanonlineapproximationtohiddenMarkovmodellearning}. 
These methods can also be used to implement world models encoding sequences of world states within planning architectures 
\citep{RueckertKappelTannebergPecevskiPeters2016RecurrentSpikingNetworksSolvePlanningTasks,BasanisiBrovelliCartoniBaldassarre2020}.
\textit{Predictive coding} is another suitable approach to implement this function \citep{RaoBallard1999Predictivecodinginthevisualcortexafunctionalinterpretationofsomeextraclassicalreceptivefieldeffects, pezzulo2014you, donnarumma2017action}.
In this respect, recent approaches integrate predictive coding with goal-oriented systems \citep{pezzulo2015active, matsumoto2022goal,jung2019goal}.


A key aspect of computation models of the GARIM theory involve the mechanisms used to support the encoding and dynamics of sub-GINPs and GINPs.
The activation of sub-GINPs could rely on local neural biased competitions taking place at different levels of abstraction (e.g., as modelled in competitive neural circuits and self-organising maps; \citealp{MysoreKothari2020MechanismsofCompetitiveSelectionaCanonicalNeuralCircuitFramework,Kohonen2001SelfOrganizingMaps,DiehlCook2015UnsupervisedLearningofDigitRecognitionUsingSpikeTimingDependentPlasticity}). The generation of GINPs could rely on local winning populations, encoding sub-GINPs, that could excite other winning populations in distal areas through long-range excitatory connections  (e.g., as modelled in \citealp{MiikkulainenBednarChoeSirosh2006ComputationalMapsintheVisualCortex}).
Neural mechanisms analogous to these have been already used in models proposed within the GNWT \citealp{DehaeneChangeux2005OngoingSpontaneousActivityControlsAccesstoConsciousnessaNeuronalModelforInattentionalBlindness}.

\subsubsection{Working memory component: the long-lasting activation of GINPs}

The working memory component should support the long-lasting activation of GINPs in the absence of their initial internal and external triggers.
\textit{Recurrent Neural Networks} (RNNs; \citealp{BarakTsodyks2014WorkingModelsofWorkingMemory}) are suitable models to emulate these functions. Indeed, they emulate the dynamic re-entrant circuits of PFC systems and basal ganglia-thalamo-cortical loops.
At the same time, basal ganglia-like selection mechanisms of the manipulator could upload/down information from such recurrent circuits (e.g., \citealp{o2006making,HolcmanTsodyks2006TheemergenceofUpandDownstatesincorticalnetworks}).
\textit{Reservoir computing} (for a review see \citealp{lukovsevivcius2009reservoir}) is another suitable approach to implement dynamic working memories. In particular, it exploits recurrent stochastic networks of which activity is `read-out' by external units. These units project back to the recurrent networks, learning to induce in them the desired dynamic pattern.
Reservoir networks are suitable to emulate different details of the brain, indeed they can be implemented with firing-rate neurons  (e.g., `echo-state networks'; \citealp{Jaeger2001Theechostateapproachtoanalysingandtrainingrecurrentneuralnetworkswithanerratumnote}) or spiking neurons (e.g., `liquid state machines'; \citealp{MaassNatschlaegerMarkram2002Realtimecomputingwithoutstablestatesanewframeworkforneuralcomputationbasedonperturbations}).

\subsubsection{Manipulator component: selection mechanisms sculpting GINPs}

The manipulator component should be able to implement the GARIM operations, thus sculpting GINPs to generate knowledge.
Computational approaches that emulate the functioning of the basal ganglia-thalamo-cortical loops could be a starting point \citep{SchrollHamker2013Computationalmodelsofbasalgangliapathwayfunctionsfocusonfunctionalneuroanatomy}. In particular, they could emulate the double-inhibition mechanisms of the basal ganglia \citep{GurneyPrescottRedgrave2001AcomputationalmodelofactionselectioninthebasalgangliaAnalysisandsimulationofbehaviour}, dynamically tuned selection processes, random exploratory selections, and focused `locking-in' selections (e.g., see \citealp{SchrollHamker2013Computationalmodelsofbasalgangliapathwayfunctionsfocusonfunctionalneuroanatomy,PrescottGonzalezGurneyHumphriesRedgrave2006ARobotModeloftheBasalGangliaBehaviorandIntrinsicProcessing,FioreSperatiMannellaMirolliGurneyFirstonDolanBaldassarre2014Keepfocussingstriataldopaminemultiplefunctionsresolvedinasinglemechanismtestedinasimulatedhumanoidrobot}).
At the same time, cortical winner-take-all processes could contribute to tune selections at finer levels (e.g.,  \citealp{MysoreKothari2020MechanismsofCompetitiveSelectionaCanonicalNeuralCircuitFramework,ArberCosta2022NetworkingBrainstemandBasalGangliaCircuitsforMovement}).
At last, lock-in mechanisms could support the prolonged activation of specific sub-GINPs (e.g., a distal goal during planning;  \citealp{BaldassarreMannellaFioreRedgraveGurneyMirolli2013IntrinsicallymotivatedactionoutcomelearningandgoalbasedactionrecallAsystemlevelbioconstrainedcomputationalmodel}).

\subsubsection{Motivation component: guiding the manipulation of GINPs}

The motivation component should guide the manipulator operations on the sub-GINPs both directly and indirectly via goals.
Low-level motivations (e.g., extrinsic motivations) could directly bias the operations of the manipulator sub-GINPs.
In addition, motivations could guide the formation/activation of goal representations, in turn guiding the manipulator to perform goal-directed manipulations (e.g., during planning or problem solving;  \citealp{SantucciBaldassarreMirolli2016GRAILaGoalDiscoveringRoboticArchitectureforIntrinsicallyMotivatedLearning,RueckertKappelTannebergPecevskiPeters2016RecurrentSpikingNetworksSolvePlanningTasks,BasanisiBrovelliCartoniBaldassarre2020,BaldassarreMannellaFioreRedgraveGurneyMirolli2013IntrinsicallymotivatedactionoutcomelearningandgoalbasedactionrecallAsystemlevelbioconstrainedcomputationalmodel}).
Motivations could also bias the acquisition of task-directed representations and not only guide their selection
\citep{GranatoCartoniDaRoldMatteraBaldassarre2021}.

Different types of motivations could play different roles.
Extrinsic motivations could be implemented with different mechanisms assigning valence to stimuli and other cognitive contents based on \textit{primary} (innate) values related to the acquisition of material resources \citep{Tye2018NeuralCircuitMotifsinValenceProcessing}. These valence should bias both selection and learning processes.
On this basis, Pavlovian associative  learning mechanisms could assign a \textit{secondary} valence to previously neutral stimuli (as done in \citealp{MannellaMirolliBaldassarre2016GoalDirectedBehaviorandInstrumentalDevaluationANeuralSystemLevelComputationalModel,mattera2020computational}). Social motivations could work on the basis of similar mechanisms but rely on social stimuli having a primary valence (e.g., see \citealp{AlfieriMatteraBaldassarre2022NeuralCircuitsUnderlyingSocialFearinRodentsanIntegrativeComputationalModel}).

Intrinsic motivations would require different mechanisms where the primary-valence stimuli originate in the system itself when it acquires knowledge and skills \citep{Baldassarre2011WhatareintrinsicmotivationsAbiologicalperspective,BaldassarreMirolli2013Intrinsicallymotivatedlearninginnaturalandartificialsystems}. 
Novelty could be supported by pattern recognition mechanisms while surprise by mechanisms based on predictors \citep{BartoMirolliBaldassarre2013Noveltyorsurprise}, similarly to what might happens in the hippocampus \citep{KumaranMaguire2007Matchmismatchprocessesunderliehumanhippocampalresponsestoassociativenovelty}.
Competence mechanisms could rely on `goal-matching processes' that compare the pursued goal with the achieved world states (\citealp{BaldassarreMannellaFioreRedgraveGurneyMirolli2013IntrinsicallymotivatedactionoutcomelearningandgoalbasedactionrecallAsystemlevelbioconstrainedcomputationalmodel}.

Emotions have more rarely been the subject of computational models \citep{MarsellaGratchPettaothers2010ComputationalModelsofEmotion}.
Models of emotional `appraisal' could be for example be used to evaluate the outcomes of internal simulations happening within the architecture
\citep{PaivaLeiteRibeiro2012Emotionmodellingforsocialrobots}.

\subsection{Towards AI systems and robotic architectures inspired by the GARIM theory}

This section illustrates the indications that the GARIM theory can provide to enhance the autonomy and effectiveness of AI and robotic systems.

\subsubsection{Adaptive functions of conscious and goal-oriented states for AI and robotic systems}

The introduction of consciousness-like  and goal-oriented processes into AI and robotic architectures could contribute to enhance several aspects of them.
The following paragraphs consider the major limitations of the current AI and robotic systems, showing how mechanisms and functions inspired by the GARIM theory might contribute to face them.

\paragraph{Flexibility}

Flexibility is still a relevant limitation of current AI systems. In particular, they are usually incapable of coping with new tasks or new conditions and to solve problems with partial knowledge
\citep{HassabisKumaranSummerfieldBotvinick2017NeuroscienceInspiredArtificialIntelligence,LakeUllmanTenenbaumGershman2017Buildingmachinesthatlearnandthinklikepeople_TargetPaper,MarcusDavis2019RebootingAIBuildingArtificialIntelligenceWeCanTrust}, although things might be changing with the most recent Large Language Models (LLM) discussed below.
The GARIM theory proposes that human behaviour flexibility depends on the brain capacity to internally manipulate the representations of goal-relevant elements (e.g., objects, goals, actions). 
These manipulations give humans the ability to actively adjust and integrate the knowledge gained in previous experiences to cope with novel goals and conditions and to improve the alignment of behaviour to goals and of these to ultimate values.
Therefore, the integration of mechanisms inspired by the GARIM theory could boost the flexibility of AI and robotic architectures.

\paragraph{Learning speed}
The learning efficiency is a second major limitation of current AI and robotic systems.
In particular, they are time consuming and need very large datasets to learn  \citep{LakeUllmanTenenbaumGershman2017Buildingmachinesthatlearnandthinklikepeople_TargetPaper,MarcusDavis2019RebootingAIBuildingArtificialIntelligenceWeCanTrust,Ullman2019UsingNeurosciencetoDevelopArtificialIntelligence}.
The GARIM theory introduces the super-ordinate representation manipulation function called \textit{Conscious Knowledge Transfer} (CKT). Based on the four GARIM operations (abstraction, specification, decomposition, composition), CKT should make the system able to transfer knowledge between tasks and domains.
In particular, the introduction of this function into AI and robotic architectures could accelerate their learning processes.
In addition, it might allow the solution of tasks with few or no direct experience on them (`few/zero shot learning', \citealp{PourpanahAbdarLuoZhouWangLimWangWu2022AReviewOfGeneralizedZeroShotLearningMethods}).

\paragraph{Creativity}

Creativity and imagination are strongly limited in AI/robotic systems \citep{HassabisKumaranSummerfieldBotvinick2017NeuroscienceInspiredArtificialIntelligence,LakeUllmanTenenbaumGershman2017Buildingmachinesthatlearnandthinklikepeople_TargetPaper,MarcusDavis2019RebootingAIBuildingArtificialIntelligenceWeCanTrust}.
The GARIM theory postulates that goal-directed top-down manipulations of perceptual and working-memory representations lead to generative and creative processes.
The development of AI/robotic architectures with these manipulation functions should boost their skills, for example making them able to elaborate creative solutions for problems.

\paragraph{Human-AI value alignment}

Many authors argue that AI systems should be able to interact safely with humans, aligning their values and goals with ours \citep{Harari2016HomoDeusaBriefHistoryofTomorrow,Bostrom2014SuperintelligencePathsDangersStrategies,Gabriel2020ArtificialIntelligenceValuesandAlignment}.
The GARIM theory provides some suggestions on how this could be done.
First, new architectures based on the GARIM theory would be more flexible, thus facilitating interactions with humans.
Mreover, they would have a motivation component, thus facilitating the design of human-like value systems \citep{Dignum2018EthicsinArtificialIntelligenceIntroductiontotheSpecialIssue}. 
In addition, they would be able to consider emotional issues, an important element to have appropriate interactions with humans \citep{HuangRustMaksimovic2019TheFeelingEconomyManagingintheNextGenerationofArtificialIntelligenceAI}.
Finally, the very function of consciousness proposed by the GARIM theory (alignment of own actions, goals and values through the manipulation of internal representations) might provide AI systems/robotic architectures with the fundamental cognitive abilities to align with human values \citep{CHRISTIANTheAlignmentProblemMachineLearningAndHumanValuesNewYorkWWNortonCompany}. 
\subsubsection{Cognitive Robotics and Machine Consciousness: designing AI and Robotics systems on the basis of the GARIM theory}
\begin{figure*}[htb!]
  \centering
     \includegraphics[width = 0.8\textwidth, height = 23 EM]{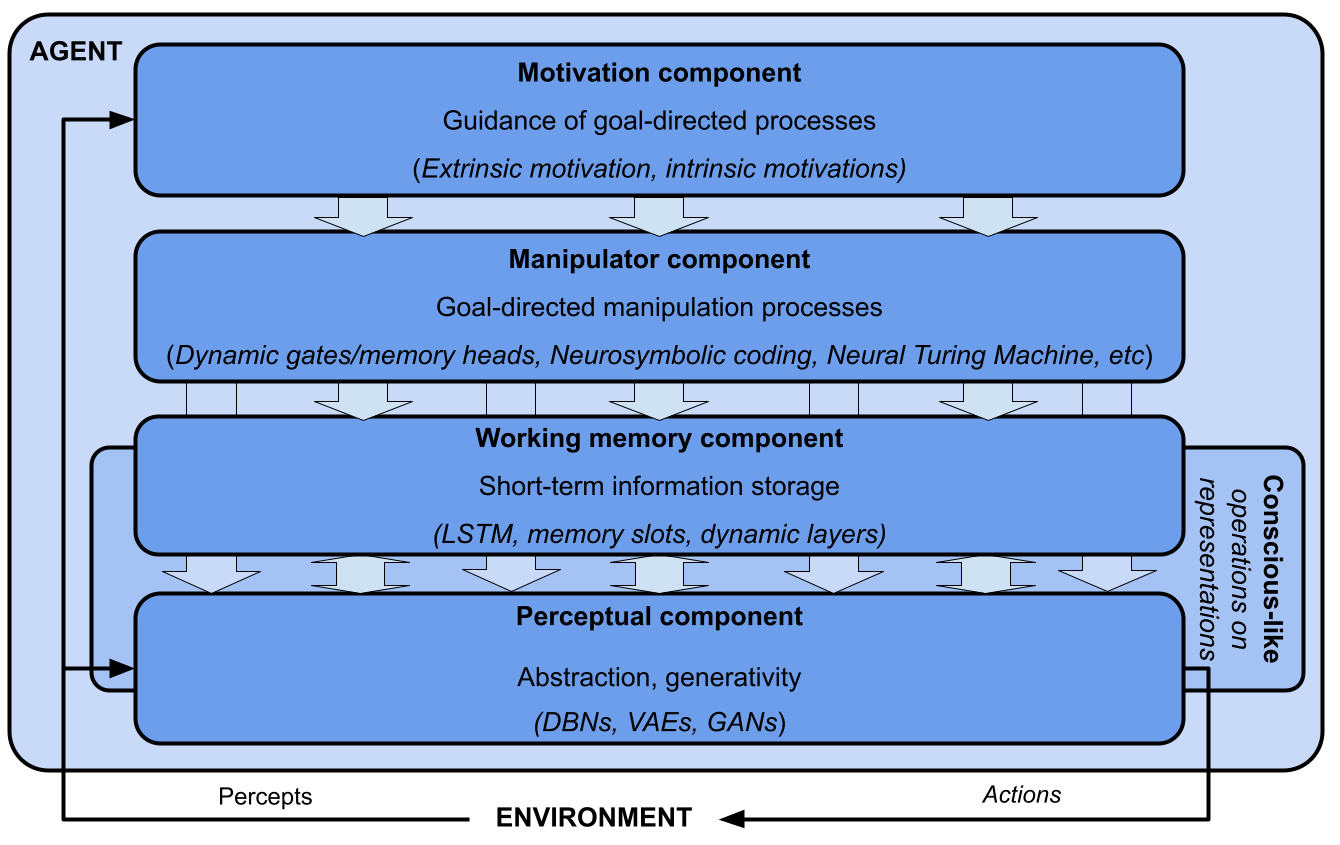}
     \vspace*{+1mm}
  \caption{
  Blueprint of an AI architecture based on the GARIM theory. 
  The figure shows some AI algorithms that could be used to implement the functions of the GARIM theory components.
  Bold text: names of the components; 
  Plain text: functions; 
  Italics text, in brackets: algorithms/models;
  Dash-highlighted text: representations on which consciousness processes operate.
  }
  \label{Figure:AIArchitecture}
\end{figure*}

This section gives initial indications on how implementing functions inspired by the GARIM theory in current AI/robotic algorithms and architectures.
Figure~\ref{Figure:AIArchitecture} illustrates a general scheme that might be followed to design specific AI and robotics systems based on the GARIM theory.

\paragraph{Perceptual component} 
This component should implement abstraction and generativity mechanisms. 
Regarding abstraction, `convolutional neural networks' (CNNs; \citealp{BengioGoodfellowCourville2017DeepLearning}) and `deep belief networks' (DBNs; \citealp{hinton2006fast,Hinton2002Trainingproductsofexpertsbyminimizingcontrastivedivergence}) are suitable approaches. Indeed, they can learn `features' of input patterns at multiple levels of abstraction.
Regarding generativity, three relevant `families' of models have been proposed \citep{BengioGoodfellowCourville2017DeepLearning}:
\textit{DBNs}, considered above, 
`variational autoencoders'
(VAEs; \citealp{KingmaWelling2013AutoEncodingVariationalBayes}), and
`generative adversarial networks' (GANs; \citealp{GoodfellowPougetAbadieMirzaXuWardeFarleyOzairCourvilleBengio2014Generativeadversarialnets}).

Although these models can be exploited to solve several tasks, they still show limitations that might prevent their use to implement manipulative functions proposed by the GARIM theory.
CNNs are not generative and are trained with a supervised algorithm.
This feature makes these networks less useful for autonomous agents.
VAEs are based on two distinct components, a bottom-up abstraction component (`encoder') and a top-down generative component (`decoder').
As a consequence, they cannot easily integrate manipulative processes because they would require two distinct manipulators.
GANs are formed by a `discriminator component' and a `generator component': the latter could be useful to implement manipulative functionalities, while the former could be used to distinguish between imagined and perceived stimuli.
Unfortunately, the `generative stochastic engine' of both VAEs and GANs is limited. 
In particular, it is located in the latent space of VAEs and in the discriminator of GANs. 
Therefore, the two networks cannot have generativity at multiple levels of abstraction.
Interestingly, DBNs show a bidirectional architecture implementing both bottom-up abstraction and top-down generative processes.
Moreover, they show two interesting features: (a) their `generative engine' is distributed into all its stochastic-units, thus supporting generativity at multiple levels of abstraction and (b) their unsupervised learning mechanisms can be integrated with reinforcement learning mechanisms, thus balancing representational requirements and task demands \citep{granato2022Integrating}.

\paragraph{Working memory component}
This component should support the information reverberation in the absence of the corresponding patterns from sensors or internal processes. 
The component should be able to learn which patterns to store and which not, also on the basis of goals.

RNNs, introduced above, are a first powerful tool usable to implement working memory.
This capacity is based on an architecture having re-entrant connections and thus capable of dynamically storing information
\citep{ChoiMatsumotoJungTani2018GeneratingGoalDirectedVisuomotorPlansBasedonLearningUsingaPredictiveCodingTypeDeepVisuomotorRecurrentNeuralNetworkModel}.
\textit{Long-short term memories} (LSTM; \citealp{HochreiterSchmidhuber1997LongShortTermMemory}) are networks based on units with a `gated self-connection' and gates in input and output connections.
The opening/closing of the gates can upload/download information in the unit, making it capable of storing memories for long times.
These networks are commonly used to solve classification and regression tasks with input sequences.
However, they have recently been updated with additional mechanisms that can support deliberative (goal-directed) processes as needed by the GARIM theory (e.g., see \citealp{JungMatsumotoTani2019GoalDirectedBehaviorunderVariationalPredictiveCodingDynamicOrganizationofVisualAttentionandWorkingMemorya}).
\textit{Neural Turing machines} \citep{GravesWayneDanihelka2014NeuralTuringMachines,WayneHungAmosMirzaAhujaGrabskaBarwinskaRaeMirowskiLeiboSantoroGemiciReynoldsHarleyAbramsonMohamedRezendeSaxtonCainHillierSilverKavukcuogluBotvinickHassabisLillicrap2018UnsupervisedPredictiveMemoryinaGoalDirectedAgent} are neural networks that support deliberative processes.
These networks use `working memory slots' that are based on numerical vectors.
These slots are read/written by `neural heads' that are trainable with gradient-based algorithms.
These features allow these networks to implement trainable logic-like reasoning. 
However, the pre-defined level of abstraction of these memory slots make them unsuitable to implement the GARIM operations of composition/decomposition, thus limiting their flexibility.

\paragraph{Manipulator component}

This component should implement two main functions. First, it should support the autonomous learning and performance of the goal-directed manipulation of representations (states, goals, actions, etc.). Second, it should support the goal-directed adaptation/tuning of these manipulation processes.
A number of AI mechanisms, introduced above, can be used to implement working memories and `neural heads', or other mechanisms, to `read/write' such memories.
These mechanisms can be important means to implement the manipulation of representations.

The implementation of goal-directed processes also requires the performance of a number of structured and temporised operations, such as the goals/sub-goals activation/de-activation.
Examples of these are: the generation and search of correct action sequences, the prediction of actions outcomes, the exchange of information between the different components of the system.
These operations are relatively easy to implement with symbolic representations and programming controls (e.g., `if-then' and `loop' operations;  \citealp{RussellNorvig2016ArtificialIntelligenceAModernApproach}) but very difficult to implement with neural mechanisms.
Current systems thus tend to be based on hybrid neural/symbolic mechanisms.
This is an important open problem as the non-neural parts of the models could obstacle the information integration capabilities of the system.
\textit{Hybrid systems }\citep{Sun2016TheCLARIONCognitiveArchitecturetowardaComprehensiveTheoryoftheMind,konidaris2018skills,oddi2019intrinsically} implement low-level cognitive processes based on neural representations and learning algorithms.
At the same time, they implement high-level cognitive processes based on symbolic representations.
This double representation format allows them, for example, to implement symbolic PDDL planning while using neural mechanisms to implement sensorimotor processes.
These approaches have limitations for our scope.
In particular, they introduce inhomogeneous representations at the low and high representation levels, requiring different mechanisms to manipulate them.
\textit{Neural Turing machines} and models like \textit{MERLIN} \citep{GravesWayneDanihelka2014NeuralTuringMachines,WayneHungAmosMirzaAhujaGrabskaBarwinskaRaeMirowskiLeiboSantoroGemiciReynoldsHarleyAbramsonMohamedRezendeSaxtonCainHillierSilverKavukcuogluBotvinickHassabisLillicrap2018UnsupervisedPredictiveMemoryinaGoalDirectedAgent} use memory slots and neural heads to perform complex tasks that require the achievement of multiple subgoals.
This approach is mainly used to solve single reactive tasks but it can also be used to solve deliberative problems \citep{ChaplotPathakMalik2021DifferentiableSpatialPlanningUsingTransformers}.
\textit{Neurosymbolic AI} (for a review see \citealp{garcez2020neurosymbolic}), and in particular recent \textit{visual planning} systems \citep{JungMatsumotoTani2019GoalDirectedBehaviorunderVariationalPredictiveCodingDynamicOrganizationofVisualAttentionandWorkingMemorya,NairPongDalalBahlLinLevine2018VisualReinforcementLearningwithImaginedGoals}, perform planning task on the basis of goal-directed processes and distributed representations (states, goals, actions, etc.).
These processes allow high flexibility, supporting generalisation capabilities that cannot be achieved by symbolic planning/problem solving. However, for now they cannot compose/decompose the manipulated elements.
\textit{Transformers}  \citep{VaswaniShazeerParmarUszkoreitJonesGomezKaiserPolosukhin2017AttentionIsAllYouNeed} implement neural internal attention mechanisms and dynamic circuits.
Their memory and attention units are integrated within the trainable input-output layers of neural networks.
These models are very effective in recalling any learnt or acquired information, even if experienced much earlier.
Transformers have been mainly used to successfully solve natural language processing tasks but require massive supervised-learning training \citep{blakeman2022selective}. 
These systems have the potential to also support deliberative processes \citep{ChaplotPathakMalik2021DifferentiableSpatialPlanningUsingTransformers} and to be hybridised with sensorimotor components \citep{DriessXiaSajjadiLynchChowdheryIchterWahidTompsonVuongYuHuangChebotarSermanetDuckworthLevineVanhouckeHausmanToussaintGreffZengMordatchFlorence2023PaLMEAnEmbodiedMultimodalLanguageModel}. In addition, they have been indicated as relevant to implement consciousness-like processes \citep{bengio2017consciousness}.
However, the functioning of AI systems based on transformers is still poorly understood \citep{AgueerayArcas2022DoLargeLanguageModelsUnderstandUs,SrivastavaEtAl2022BeyondTheImitationGameQuantifyingAndExtrapolatingTheCapabilitiesOfLanguageModels}.
In this respect, in the future it might be interesting to evaluate if manipulation operations similar to those proposed by the GARIM theory actually take place within AI systems based on transformers.

\paragraph{Motivation component}

Extrinsic motivations are usually emulated trough reward signals \citep{sutton1998introduction}.
Moreover, `pseudo-rewards' can be used to guide model-based hierarchical reinforcement learning based on goal-matching events \citep{BotvinickNivBarto2008HierarchicallyOrganizedBehaviorandItsNeuralFoundationsaReinforcementLearningPerspective}.

Intrinsic motivations have demonstrated to effectively support the autonomous acquisition of knowledge of robots \citep{BaldassarreMirolli2013Intrinsicallymotivatedlearninginnaturalandartificialsystems}. 
Indeed, intrinsic motivation mechanisms can drive the investigation and learning of novel/surprising experiences, leading to the acquisition of new state representations and models \citep{schmidhuber1991possibility,OudeyerKaplanHafner2007IntrinsicMotivationSystemsforAutonomousMentalDevelopment,BartoMirolliBaldassarre2013Noveltyorsurprise,CartoniBaldassarre2018Autonomousdiscoveryofthegoalspacetolearnaparameterizedskill}. 
Moreover, they can lead to the acquisition of `intrinsic goals' (autonomously found) and motor skills to accomplish them \citep{BartoSinghChentanez2004IntrinsicallyMotivatedLearningofHierarchicalCollectionsofSkills,SantucciBaldassarreMirolli2016GRAILaGoalDiscoveringRoboticArchitectureforIntrinsicallyMotivatedLearning,NairPongDalalBahlLinLevine2018VisualReinforcementLearningwithImaginedGoals}.
Intrinsic motivations are commonly used to guide intelligent machines and robots to seek knowledge in the \textit{external} environment.
Instead, according to the GARIM proposal they could guide the \textit{internal} building of the knowledge that the agent lacks.

There are few AI approaches that emulate the generation of emotions (for reviews see \citealp{PaivaLeiteRibeiro2012Emotionmodellingforsocialrobots,mirolli2010roles,SunWilsonLynch2016EmotionaUnifiedMechanisticInterpretationfromaCognitiveArchitecture}).
These models could be used as a starting point for implementing emotion-based evaluation of internal representations.

\paragraph{Open challenges: what is missing?}

The two architecture schemes we proposed for guiding the development of computational models (Figure~\ref{Figure:RIMBlueprintArchitecture}) and AI/robotic architectures (Figure~\ref{Figure:AIArchitecture}) include the main elements that should support conscious and flexible artificial systems.
For example, they include the main features of machine consciousness systems (self-modelling, information broadcasting, higher-level representations, attention processes, and information integration; \citealp{reggia2013rise}).
Moreover, they include the fundamentals axioms of Machine Consciousness (world models, imagination, attention, planning, and affective evaluation; \citealp{aleksander1995artificial}) .
However, critical elements for building conscious machines may still be missing.

First, the four macro-systems proposed by the GARIM theory require important low-level functions to support the emergence of GINPs.
For example, the brain shows a high capacity for generating \textit{associations} and avoiding unbounded activations. These capacities are based on grid-like circuits and finely regulated inhibitory processes.
These features are missing in common artificial neural network architectures, which favour bottom-up/top-down directional information flows with few recurrences \citep{LynnBassett2019ThePhysicsofBrainNetworkStructureFunctionandControl}.
Second, the brain exhibits highly dynamic processes that could be based on fixed-point/cycle/strange attractors. ANNs are still not able to fully emulate these processes. These elements might be needed to implement the GARIM operations on sub-GINPs \citep{Breakspear2017DynamicModelsofLargeScaleBrainActivity}.
Third, the flexible selection functions implemented by the basal ganglia-thalamo-cortical loops are only partially captured by current neural systems.
Fourth, strongly-coupled sensorimotor loops engaged by animals with the environment are often absent in AI systems.
Moreover, current robots have still a very limited autonomy to interact with the environment.
Agent-environment interactions might instead be very important to acquire internal representations strongly coupled with the real environment.
Last, till recently AI/robotic systems lacked the capacity to suitably integrate language with sensorimotor experience, in particular they lacked a meaning and understanding grounded on sensorimotor experience.
Recently, however, large language models have been argued to acquire some meaning even without sensorimotor grounding 
\citep{2021DoesVisionandLanguagePretrainingImproveLexicalGrounding,AbdouKulmizevHershcovichFrankPavlickSoegaard2021CanLanguageModelsEncodePerceptualStructureWithoutGroundingACaseStudyInColor,AgueerayArcas2022DoLargeLanguageModelsUnderstandUs,SrivastavaEtAl2022BeyondTheImitationGameQuantifyingAndExtrapolatingTheCapabilitiesOfLanguageModels}, and grounding might be realised very soon with systems integrating language and sensorimotor capabilities \citep{DriessXiaSajjadiLynchChowdheryIchterWahidTompsonVuongYuHuangChebotarSermanetDuckworthLevineVanhouckeHausmanToussaintGreffZengMordatchFlorence2023PaLMEAnEmbodiedMultimodalLanguageModel}. 
The fact that these models are based on transformers, which might perform operations very similar to the representation manipulations postulated by the GARIM theory, hence represents an interesting topic for future investigations.

Overall, the realisation and integration of all these elements is still a great open challenge.
Much of the flexibility of the brain is based on its highly structured and integrated architecture, which seems difficult to reproduce in artificial systems.
Indeed, the brain integrates habitual and goal-directed processes and it is the product of a long evolutionary process that is hard to reproduce in machines  \citep{BaldassarreSantucciCartoniCaligiore2017ThearchitecturechallengeFutureartificialintelligencesystemswillrequiresophisticatedarchitecturesandknowledgeofthebrainmightguidetheirconstruction,baldassarre2020goal,CaligioreArbibMiallBaldassarre2019TheSuperLearningHypothesisIntegratingLearningProcessesacrossCortexCerebellumandBasalGanglia,Ullman2019UsingNeurosciencetoDevelopArtificialIntelligence}.
Producing conscious intelligent machines without relying on such a highly integrated architecture is therefore a great challenge.

\section{Conclusions}

In this work we introduce the \textit{Goal-Aligning Representation Internal Manipulation} (GARIM) theory of flexible goal-directed cognition and consciousness. 
The central idea of the GARIM theory is that conscious states support the active manipulation of internal representations, making them more aligned with the goals pursued. This goal-oriented alignment leads to the generation of the necessary knowledge to face novel situations and goals and to make goal-directed behaviour more flexible and effective. 
The GARIM theory postulates that a conscious goal-directed behaviour is characterised by five distinctive elements.
First, consciousness serves the adaptation of goal-directed behaviours. In particular, consciousness processes support the goal-aligning manipulation of internal representations, in turn boosting the flexibility of goal-directed behaviour. 
Second, the theory hypotheses the existence of `Goal-based Integrated Neural Patterns' (GINPs). These are distributed active neural representations that (a) are consciously perceived and thus intentionally manipulable (consciousness level), and (b) are closely related the pursued goals (goal-relevance). Different levels of these two dimensions lead to representations characterised by different levels of consciousness (Non-GINPs, Pre-GINPs, Temp-GINPs, GINPs)  and representational proprieties (e.g., information integration).
Third, the GARIM theory specifies that goal-directed manipulations rely on four key `components', namely four partially overlapping anatomo-functional brain macro systems (perceptual working-memory, abstract working-memory, internal manipulator, motivational component).
Fourth, previous systems give rise to four classes of computational operations (GARIM operations) that support  the representation manipulations (abstraction, specification, decomposition, composition). 
Fifth and last, the GARIM theory introduces the concept of `GARIM agency', a sense of agency that emerges from conscious goal-directed processes and, in particular, representation manipulations. These manipulations lead to the generation of a subjective internal reality supported by three key features: self-model, 
emotional/perceptual vividness, and 
manipulation control. 
On the basis of the GARIM agency and its key features, the theory proposes different levels of conscious states (from reactive behaviours to highly flexible goal-directed 
behaviours).
In addiction to clarifying neuro-computational processes at the basis of conscious and flexible goal-directed behaviours, the GARIM theory has both scientific and technological implications. For example, it clarifies some aspects of subjective experience and agency, also introducing a potential quantitative scale. Moreover, it accounts for several elements of current theories of consciousness, integrating them into a common functional and computational framework that focuses on goal-directed processes. The GARIM theory also generates insights for experimental and clinical fields. In particular, it proposes clinical insights, experimental predictions and new ideas for building experimental protocol of goal-directed behaviour and consciousness. At last, the GARIM theory furnishes indications for building new computational models, and AI/robotic architectures. In particular, it proposes that conscious goal-aligning  manipulations of representations could enable  AI/robotic architectures to achieve human-like flexibility and general intelligence.

\section{Acknowledgements}
This work has received funding from the European Union’s Horizon 2020 Research and Innovation Program with the projects
`GOAL-Robots -- Goal-based Open-ended Autonomous Learning Robots', GA N. 713010,
`HBP -- Human Brain Project SGA3', GA N. 945539;
and from the Horizon Europe Program with projects 
`PILLAR-Robots - Purposeful Intrinsically motivated Lifelong Learning Autonomous Robots', GA N. 101070381, 
and `EBRAINS-Italy - European Brain ReseArch INfrastructureS Italy', PNRR N. IR000001, CUP B51E22000150006. 
We thank Emilio Cartoni and Andrea Mattera for the useful feedback on the early versions of the manuscript.

\bibliography{MyBibliography.bib}
\end{document}